\documentclass[letterpaper]{article} 
\usepackage{aaai24}
\usepackage{times}  
\usepackage{helvet}  
\usepackage{courier}  
\usepackage[hyphens]{url}  
\usepackage{graphicx} 
\usepackage{natbib}  
\usepackage{caption} 
\usepackage{algorithm}
\usepackage{algorithmic}

\usepackage{newfloat}
\usepackage{listings}
\DeclareCaptionStyle{ruled}{labelfont=normalfont,labelsep=colon,strut=off} 
\floatstyle{ruled}
\newfloat{listing}{tb}{lst}{}
\floatname{listing}{Listing}
\title{DSD$^{\boldsymbol{2}}$: Can We Dodge Sparse Double Descent and Compress the Neural Network Worry-Free?}
\author {
    Victor Qu\'etu\textsuperscript{},
    Enzo Tartaglione\textsuperscript{}
}
\affiliations {
    \textsuperscript{}LTCI, T\'el\'ecom Paris, Institut Polytechnique de Paris, France\\
    \{name.surname\}@telecom-paris.fr
}

\usepackage{bibentry}

\usepackage{graphicx}
\usepackage{subcaption}
\usepackage{amsmath}
\usepackage{amssymb}
\usepackage{mathtools}
\usepackage{dsfont}
\usepackage{algorithm}
\usepackage{algorithmic}
\usepackage{multirow}
\usepackage{enumitem}
\usepackage{tikz}
\usepackage{booktabs} 
\newcommand\tikzmark[1]{\tikz[remember picture] \node (#1) {};}

\newtheorem{obs}{Observation}
\usepackage{soul}

\newcommand{\alglinelabel}{%
  \addtocounter{ALC@line}{-1}
  \refstepcounter{ALC@line}
  \label
}

\begin{document}

\maketitle

\begin{abstract}
Neoteric works have shown that modern deep learning models can exhibit a sparse double descent phenomenon. Indeed, as the sparsity of the model increases, the test performance first worsens since the model is overfitting the training data; then, the overfitting reduces, leading to an improvement in performance, and finally, the model begins to forget critical information, resulting in underfitting. Such a behavior prevents using traditional early stop criteria.

In this work, we have three key contributions. First, we propose a learning framework that avoids such a phenomenon and improves generalization. Second, we introduce an entropy measure providing more insights into the insurgence of this phenomenon and enabling the use of traditional stop criteria. Third, we provide a comprehensive quantitative analysis of contingent factors such as re-initialization methods, model width and depth, and dataset noise. The contributions are supported by empirical evidence in typical setups. Our code is available at \url{https://github.com/VGCQ/DSD2}.
\end{abstract}


\section{Introduction}
\label{sec: Introduction}
\let\svthefootnote\thefootnote
\newcommand\freefootnote[1]{%
  \let\thefootnote\relax%
  \footnotetext{#1}%
  \let\thefootnote\svthefootnote%
}
\begin{figure}[t]
    \centering
    \includegraphics[width=\columnwidth]{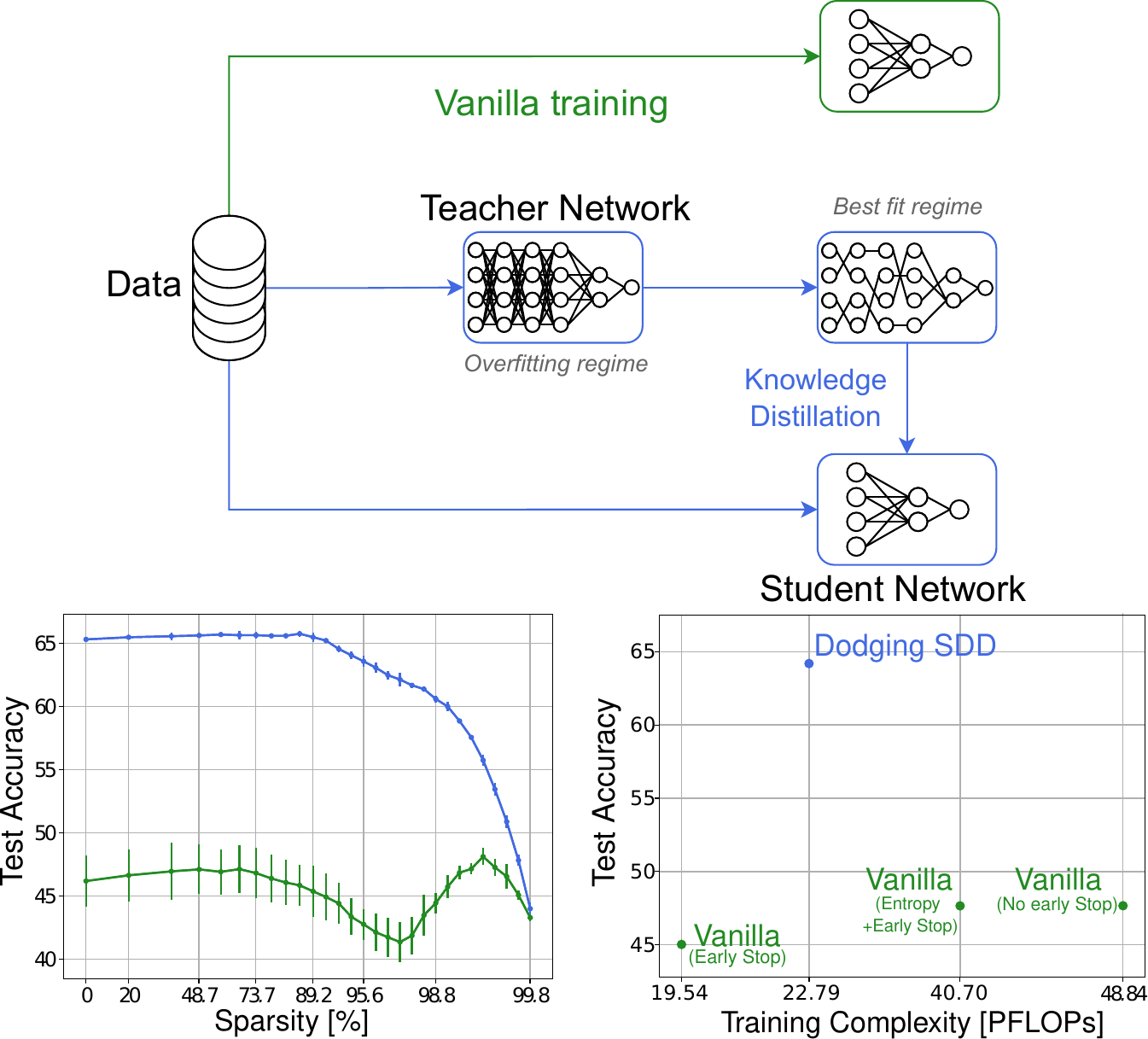}
    \caption{Distilling knowledge from a sparse teacher grants access to solutions (for the student model) where SDD is dodged, also saving computation.}
    \label{fig:Teaser_fig}
\end{figure}
\freefootnote{This paper has been accepted for publication at the 38th Annual AAAI Conference on Artificial Intelligence (AAAI24).}Nowadays, deep neural networks are one of the most employed algorithms when required to solve complex tasks. In particular, their generalization capability allowed them to establish new state-of-the-art performance in domains like computer vision~\cite{he2016deep, dosovitskiy2021an} and natural language processing~\cite{vaswani2017attention, brown2020language}, showing as well promising capability in very complex hybrid tasks, like text-to-image generation~\cite{ramesh2022hierarchical, saharia2022photorealistic}. The problem of optimally sizing these models is relevant to the vastly distributed employment of deep neural networks on edge devices~\cite{chen2019deep, lin2022ondevice}, posing questions about power consumption and hardware complexity~\cite{goel2020survey, luo2022lightnas}.

It was general knowledge that the more a model is overparametrized, the easier it will overfit the training set, entering the \emph{memorization phase}: the model memorizes the single samples in the training set, learning also a wrong set of features. Such a phenomenon harms the model's generalization, worsening its performance on unseen data~\cite{liu2020early}. Recently, a new surprising phenomenon, \emph{double descent}~(DD)~\cite{belkin2019reconciling}, has been observed for extremely over-parametrized models: beyond the traditional over-fitting regime, while continuing to increase the size of the model, the generalization gap between train and test performance inverts trend, and narrows the more, the larger the model is~\cite{nakkiran2021deep}.

The DD phenomenon raises questions about how to optimally size the model to have the best performance (while having the minimum size). Many approaches have indeed proposed the use of regularization functions to relieve DD in models for regression and classification tasks. However, when moving to real applications, the complexity of optimally tuning the regularization hyper-parameters and learning optimal early-stop discourages their use~\cite{kan2020avoiding}. While DD's typical analysis concerns moving from small models to big ones, a recent work observed a similar phenomenon also moving from an over-parametrized model backward to a smaller one~\cite{SparseDoubleDescent}. That is possible thanks to pruning, which iteratively removes parameters from the model. Intuitively, while pruning removes parameters from the model, while at first the performance is enhanced, it will enter a second phase where its inevitable worsening is met~\cite{han2015learning,10222624}. Such effect took the name of \emph{sparse double descent} (SDD): is it inevitable? Can we properly regularize the model toward test performance enhancement? What is the underlying explanation of the SDD phenomenon?

\noindent We summarize our contributions as follows.
\begin{itemize}[noitemsep, nolistsep]
    \item To the best of our knowledge, we propose the first approach avoiding SDD, and providing a model with a good performance in terms of validation/test accuracy, consistently. More specifically, we regularize a student model distilling knowledge from a sparse teacher (in its best validation accuracy region), observing that
the student dodges SDD (Fig.~\ref{fig:Teaser_fig}). Interestingly, we observe that this happens even when distilling knowledge from a non-pruned teacher.
    \item We study SDD from the perspective of ``neuron's states'': 
we calculate the entropy of the activations in the model,  
observing a correlation between the \emph{interpolation regime} (where the SDD occurs) and the entropy's flatness. When leaving such a region, thus entering the \emph{classical regime} where the traditional bias/variance trade-off occurs,  the entropy monotonically decreases: a check on this measure enables-back the use of early stop criteria, which saves training computational cost (rightmost Fig.~\ref{fig:Teaser_fig}). 
    \item We propose a quantitative study on some open questions, in particular: i) is DD/SDD still occurring when models increase in depth? ii) do we find the best validation/test performance configuration in models extremely over-parametrized or right before under-fitting? iii) is there a big difference between setting back parameters to their initial value (rewinding), randomly initializing, and not perturbing the model's parameters after pruning?
\end{itemize}

\begin{figure*}[t]
    \begin{subfigure}{0.5\textwidth}
        \centering
        \includegraphics[width=0.95\textwidth]{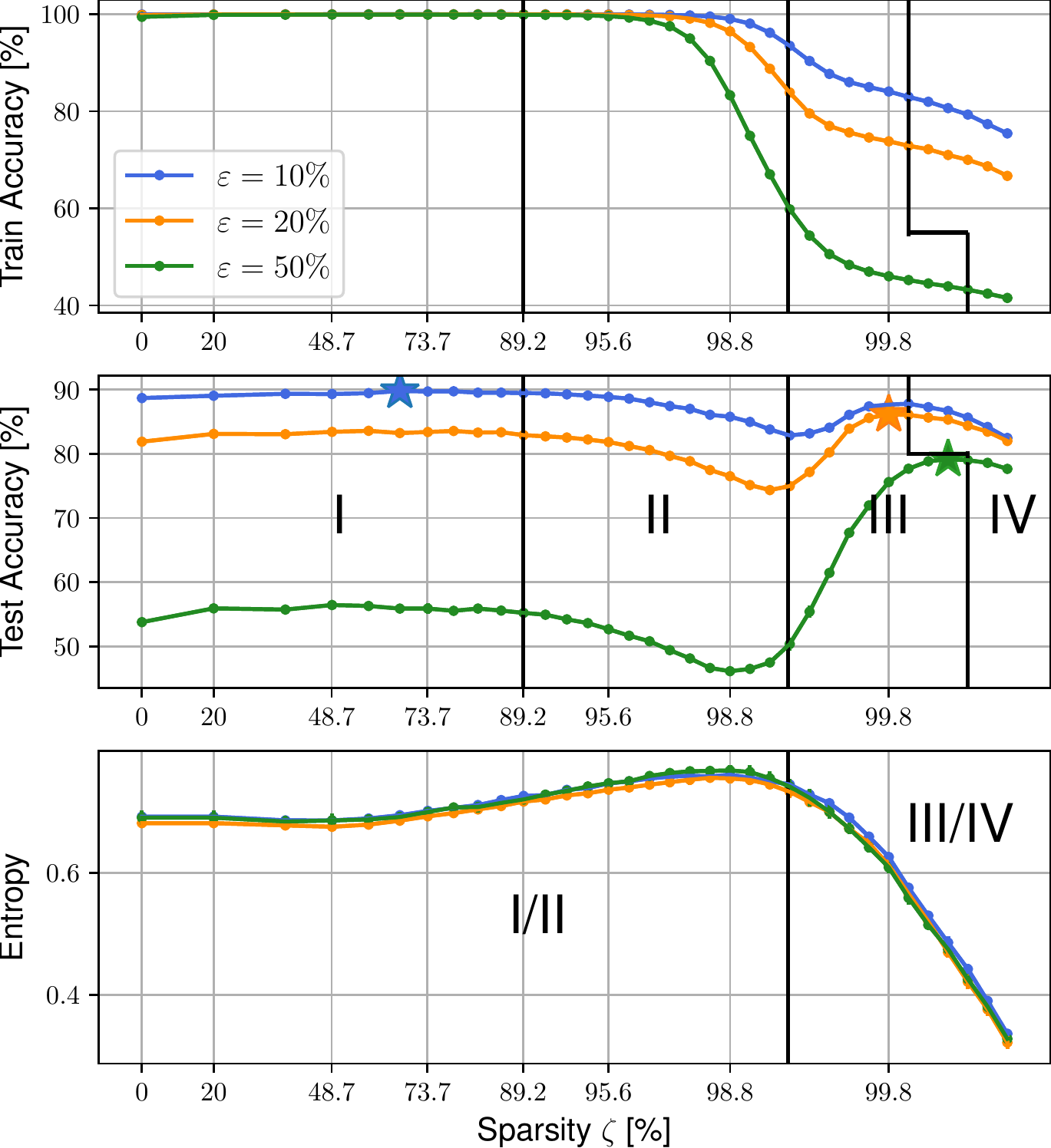}
        \caption{~}
        \label{fig:Teachers_CIFAR-10}
    \end{subfigure}
    \begin{subfigure}{0.5\textwidth}
        \centering
        \includegraphics[width=0.95\textwidth]{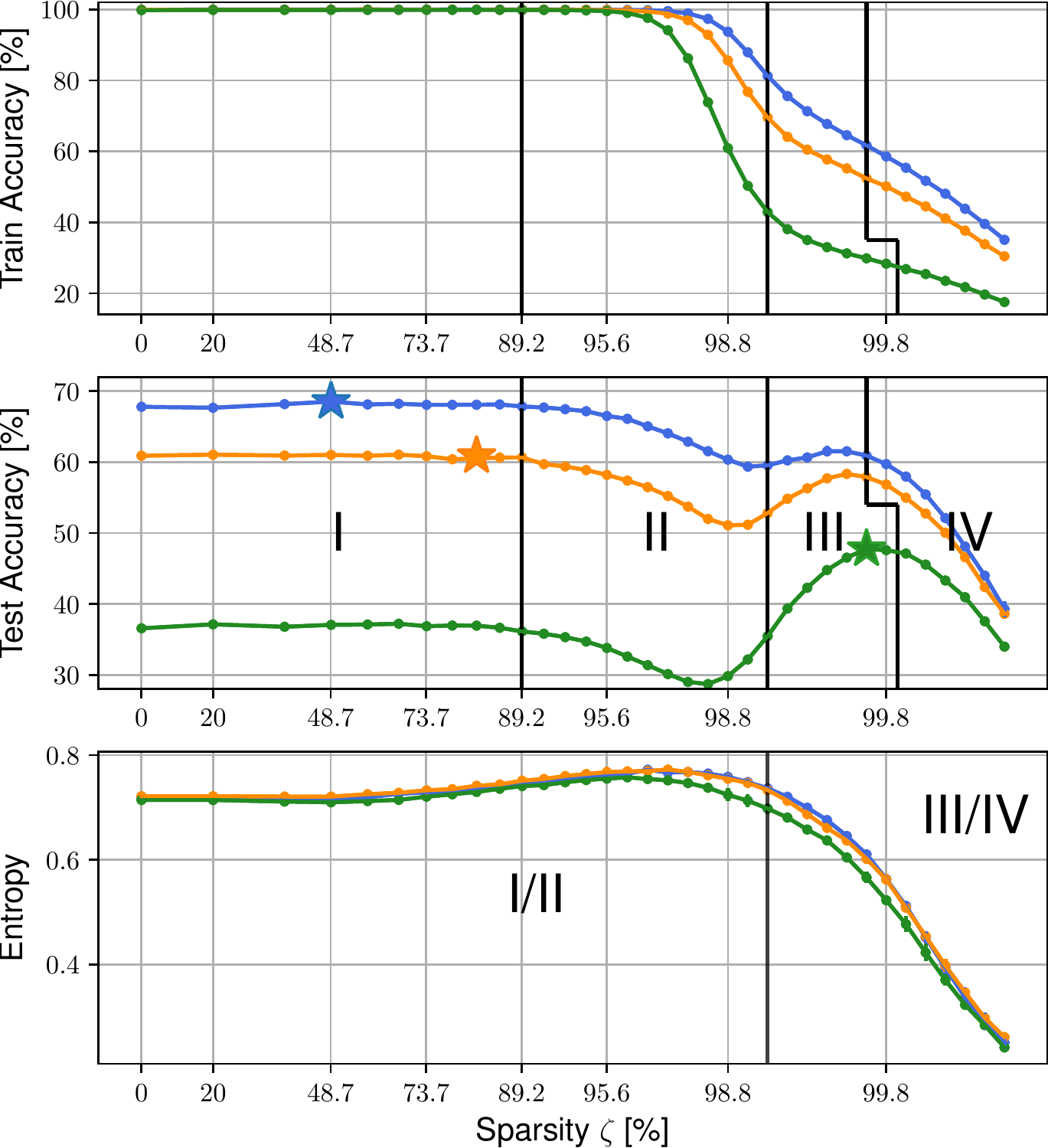}
        \caption{~}
        \label{fig:Teachers_CIFAR-100}
    \end{subfigure}
    \vspace{-15pt}
    \caption{Performance of ResNet-18 with different amount of noise $\varepsilon$ on CIFAR-10~(a) and CIFAR-100~(b). \textbf{I}: Light Phase. \\\textbf{II}: Critical Phase. \textbf{III}: Sweet Phase. \textbf{IV}: Collapsed Phase.}
    \label{fig:phases}
\end{figure*}

\section{Related works}
\label{sec: Related works}

\paragraph{The real world is noisy} In the real world, data acquisition is often noisy~\cite{gupta2019dealing}, stemming from data collection or labeling. Concerning the annotation noise, many works proposed solutions to prevent the learning of wrong feature sets: for example, \cite{li2017learning} propose a unified distillation framework by leveraging the knowledge learned from a small clean dataset and semantic knowledge graph to correct the noisy labels. Other works find solutions inspired by the benefits of noise in the nervous system: \cite{arani2021noise}, for example, show that injecting constructive noise at different levels in the collaborative learning framework enables training the model effectively and distills desirable characteristics in the student model. More specifically, they propose methods to minimize the performance gap between a compact and a large model, to train high-performance compact adversarially-robust models. 

Since a single image may belong to several categories, different samples can suffer from varying intensities of label noise. \cite{xu2020feature} proposed a simple yet effective feature normalized knowledge distillation that introduces the sample-specific correction factor to replace the temperature. \cite{kaiser2022blind} developed a teacher-student approach that identifies the tipping point between good generalization and overfits, thus estimating the noise in the training data with Otsu's algorithm. Other works focus more on the label's prediction robustness: \cite{sau2016deep}, for example, introduced a simple method that helps the student to learn better and produces results closer to the teacher network by injecting noise and perturbing the logit outputs of the teacher. With this setup, the noise simulates a multi-teacher setting and produces the effect of a regularizer due to its presence in the loss layer. To simulate noise in the labels, a typical approach is to manually inject noise in some well-annotated, standard datasets like MNIST and CIFAR-10/100~\cite{nakkiran2021deep, SparseDoubleDescent}. AI security works use a similar setup as well, where noise is injected parametrically to analyze the model's robustness against attacks. In a nutshell, these attacks propose adversarial representations of the data and check the model's performance - so this setup is named ``adversarial learning''~\cite{9013065}. \\

\paragraph{Double Descent in classification tasks.}
Considering the presence of labeling noise, the occurrence of DD is a real threat. DD has already been reported in various machine learning models, like decision trees, random features~\cite{meng2022multiple}, linear regression~\cite{muthukumar2020harmless}\cite{belkin2020two, hastie2022surprises}, and deep neural networks~\cite{yilmaz2022regularization}. For classification tasks, the test error of standard deep networks, like the ResNet architecture, trained on image classification datasets, consistently follows a double descent curve both when label noise is injected (CIFAR-10), and in some cases, even without any label noise injection (CIFAR-100)~\cite{yilmaz2022regularization}. \cite{nakkiran2021deep} show that double descent occurs not just as a function of model size when increasing the model width, but also as a function of the number of training epochs. The double descent phenomenon has been extensively studied under the spectrum of over-parametrization~\cite{nakkiran2021deep, chang2021provable}.\\
\cite{SparseDoubleDescent} observe the occurrence of double descent not only in the traditional setups but also when un-structurally pruning a dense model, observing the SDD phenomenon. Working on a related research question, and motivating the importance of further studying the SDD, \cite{chang2021provable} addressed the important question of whether it could be more convenient to train a small model directly, or quite first train a larger one and then prune it. In this work, the authors provide convincing evidence that the latter strategy is winning, toward enhanced model performance in sparsified regimes. \citet{cotter2021distilling}, as a contrast to this work's purpose, exploited the DD phenomenon in a self-supervised setup, to assign pseudo-labels to a large held-out dataset.
While this work sought to exploit DD, our goal is different: our objective is to avoid the sparse double descent, towards enhanced performance on the final student model, on the same task and dataset as the teacher. In the next section, we present and show the occurrence of SDD.

\section{Model size and sparse double descent}
\label{sec : Model size reduction}
\subsection{Background on neural network's pruning}
Neural network pruning aims to reduce a large network while maintaining accuracy by removing irrelevant weights, filters, or other structures, from neural networks. All the pruning algorithms removing weights without explicitly considering the neural network's structure are typically named \emph{unstructured} pruning methods.
Various unstructured pruning methods exist and can be divided into magnitude-based, where the ranking for the parameters to prune is based on their magnitude~\cite{han2015learning, louizos2018learning, h.2018to}, and gradient-based, where the ranking or the penalty term is a function of the gradient magnitude (or to higher order derivatives)~\cite{lee2018snip, tartaglione2022loss}. The general comparison between the effectiveness of any of the reported approaches is reported by~\cite{blalock2020state} and, although complex pruning approaches exist, the simple magnitude-based one, in general, is considered a good trade-off between complexity and competitiveness~\cite{Gale_Magnitude}: hence, we will focus on this one.\footnote{\cite{SparseDoubleDescent} showed that magnitude, gradient-based, and random pruning achieve similar performance for the same setup as we consider in this work. Moreover, we also present a study employing structured $\ell_1$-pruning in Appendix.}

Training first an over-parametrized model, and then pruning it, leads to an improved generalization. In particular, \cite{chang2021provable} analyzes the beneficial effects of pruning, and then compares the performance achieved by pruned models to shallow vanilla ones. This work motivates the quest for investigating SDD, looking for the highest possible performance on unseen data.

\subsection{Pruning exhibits sparse double descent}
\paragraph{Setup} The trained model $\mathcal{M}$ on the train set $\mathcal{D}_{\text{train}}$ (whose performance is evaluated on the validation set $\mathcal{D}_{\text{val}}$) consists of $L$ layers, having $\boldsymbol{w}^\mathcal{M}$ as its set of parameters, and $\boldsymbol{w}_l^\mathcal{M}$ indicates those belonging to the $l$-th layer. When we prune the $\zeta$-th fraction of parameters from the model, the parameters are projected to a parameter sub-space, according to a threshold on the quantile function
$\mathcal{Q}_{\mathcal{M}}(\cdot)$, computed on the absolute values for parameters in $\mathcal{M}$. 

The overall approach employed to reduce the dimensionality of the trained $\mathcal{M}$ follows these steps. The first step is to train the dense model. Until it has reached the desired sparsity percentage $\zeta_{\text{wall}}$, the model is iteratively pruned using some pruning strategy (i.e. magnitude pruning, following~\cite{SparseDoubleDescent}), perturbed (weights can be rewound to initialization, randomly re-initialized, or not perturbed at all), and the sparse model is re-trained on $\mathcal{D}_{\text{train}}$. The training follows standard policies: the set of hyper-parameters as well as the algorithm are reported in Appendix. 
When $\zeta_{\text{wall}}$ is reached, the model parameters $\boldsymbol{w_{\text{best}}}$, which achieve the best performance on the validation set $\mathcal{D}_\text{val}$ are returned.

\paragraph{Experiments} As in~\cite{SparseDoubleDescent}, SDD is, consistently, found in our experiments. In particular, Fig.~\ref{fig:Teachers_CIFAR-10} and Fig.~\ref{fig:Teachers_CIFAR-100} show double descent of a ResNet-18 model trained on CIFAR-10 and CIFAR-100, respectively. The four phases reported by~\cite{SparseDoubleDescent} are displayed in Fig.~\ref{fig:phases} (first a \emph{light} phase, secondly a \emph{critical} phase, thirdly the \emph{sweet} phase and finally, the \emph{collapsed} phase).

\subsection{Better low parametrization or extreme over parametrization?}
\label{sec:earlystop}
Currently, there is a big debate about whether some simple techniques, like early stopping, are sufficient to achieve great generalization performance. In particular, \cite{Rice2020OverfittingIA} studies the extreme over-parametrization for adversarially trained deep networks. The authors observe that overfitting the training set harms robust performance to a large degree in adversarially robust training across multiple datasets: this can be fought by simply using early stopping. To mitigate the robust overfitting, \cite{Chen2021RobustOM} propose self-training to smoothen the logits, combined with stochastic weight averaging trained by the same model, the other performing stochastic weight averaging~\cite{izmailov2018averaging}. Although these works focus on DD, similar effects can be drawn to SDD: is it always true that the best model is in the sweet phase?

In our results presented in Fig.~\ref{fig:phases}, we observe a correlation between the best model (marked with $\star$) and the noise in the dataset: on CIFAR-10, $\boldsymbol{w_{\text{best}}}$ is located in the sweet phase for $\varepsilon \in \left\{20\%, 50\%\right\}$, while on CIFAR-100, $\boldsymbol{w_{\text{best}}}$ is in the sweet phase only for $\varepsilon=50\%$. Therefore, we empirically observe a correlation between the amount of noise in the training set and the location of $\boldsymbol{w_{\text{best}}}$: for small noise (i.e.~$<15\%$), $\boldsymbol{w_{\text{best}}}$ is located in the Light Phase (I). As $\varepsilon$ exceeds $15\%$, $\boldsymbol{w_{\text{best}}}$ is consistently found in the Sweet Phase (III). A further analysis of this shift is conducted in Appendix.

\subsection{Critical phase occurrence}
\label{sec:hessian}
\begin{figure}[t]
    \centering
    \includegraphics[width=0.95\linewidth]{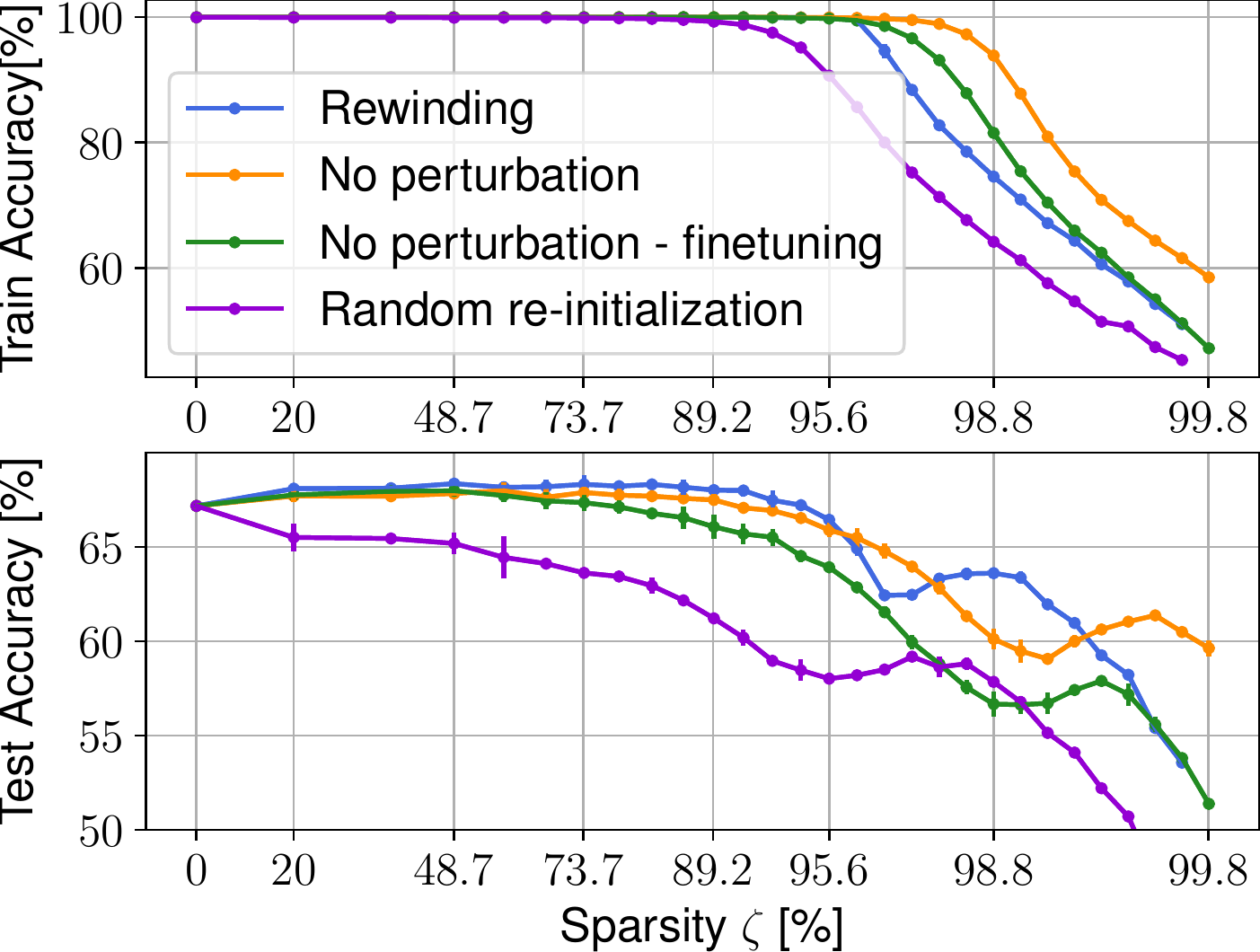}
    \caption{Performance of ResNet-18 on CIFAR-100 with $\varepsilon=10\%$ when retrained from either the original initialization (lottery ticket), a random re-initialization, or from the last configuration achieved before pruning.}
    \label{fig:compare_perturbation}
\end{figure}

\cite{frankle2018lottery} conjectured, under the so-called \emph{lottery ticket hypothesis}, that a large network contains smaller sub-networks, which can be trained in isolation without performance loss. Their approach consists of rewinding the parameters to their original value every pruning iteration and is one of the most popular for neural network pruning~\cite{malach2020proving, zhang2021efficient}. 

We compare, in Fig.~\ref{fig:compare_perturbation}, different strategies to introduce perturbations in the model's parameters - random re-initialization, rewinding (in the lottery ticket hypothesis fashion), and not introducing any perturbation-, averaged on three seeds. We observe, in all the above-mentioned cases, the rise of the sparse double descent. However, we notice a gap in the performance between the different perturbation approaches. While randomly re-initializing the model leads to worse results, rewinding and not introducing any perturbation exhibit slightly different behavior, depending on the pruning regime. In particular, rewinding can marginally achieve a better performance, but the performance decays faster than not-perturbing. Moreover, we investigate whether it could be more convenient to let the model remain in the neighborhood of the same local minimum found by the dense model. Towards this end, we propose an experiment where we do not perturb the learned parameters, and we scale down the learning rate. In this scenario, we observe the performance consistently deteriorated.
Our finding suggests that with rewinding we are post-posing, in the $\zeta$ plane, the critical phase, and without introducing any perturbation we are further post-posing it: we will use a ``no perturbation'' scheme for all our experiments.

\subsection{An entropy-based interpretation to the sparse double descent}
\label{sec:entropy}
Analyzing the SDD phenomenon from a learning dynamics perspective raises questions related to the so-called \emph{information bottleneck theory}~\cite{tishby1999proceedings, tishby2015deep}. In particular, this approach estimates the mutual information between the information processed by the layers and the input and output variables. Hence, it is possible to calculate optimal theoretical limits and set the bars for the generalization error. Several works have followed this theory, observing differences in the learning dynamics for different activation functions employed~\cite{michael2018on}, improving the estimation approach~\cite{pan2021disentangled}, and verifying that such theory admits DD in regression tasks~\cite{ngampruetikorn2022information}. Inspired by these works, we formulate the following observation.
\begin{obs}
    As the size of the model, in terms of the number of parameters per neuron, shrinks from the light phase, the entropy of the features inside the model is stationary (as small adjustments in the parameters are needed). As we leave the interpolation regime right after the interpolation threshold, the entropy begins to decrease. 
\end{obs}
To empirically verify this observation, we can visualize the entropy of the activations in the model. Considering that our observation is not constrained to any validation/test set, we will perform the measures directly on the training set. We define the average neuron's entropy in the $l$-th layer as
\begin{equation}
    \label{eq:entr}
    \bar{\mathcal{H}}_{l|\mathcal{D}_{\text{train}}} = - \frac{1}{N_l}\sum_{i=1}^{N_l} \sum_{\zeta\in \{0;1\}}p\left(s_{l,i}^{\zeta}\right) \log\left[p\left(s_{l,i}^{\zeta}\right)\right],
\end{equation}
where $p\left(s_{l,i}^{\zeta}\right)$ is the probability (in a frequentism sense, over $\mathcal{D}_{\text{train}}$) the $i$-th neuron (ReLU-activated) in the $l$-th layer to be in the negative region ($\zeta=0$) or in the positive one ($\zeta=1$), similarly to how done in~\cite{liao2023can, spadaro2023shannon}.
We provide more details on the entropy computation in Appendix. 
Fig.~\ref{fig:phases} displays the variation of the entropy, averaged across all the model's layers. For the considered examples, we corroborate our observation, enabling-back the use of early-stop criteria jointly with the entropy. Indeed, the entropy stays stationary and then decreases when the model enters the classical regime (entering the sweet phase).  Using traditional early criteria starting from this regime saves training computation as the pruning/training process is stopped when the performance decreases. We provide more details in Appendix.

\subsection{Generalization gap in deep double descent: relationships between DD and SDD}
\label{subsec : Relationships between DD and sparse DD}

In this section, we analyze the occurrence of the sparse double descent concerning the model's size. To carry out the following experiments, we have defined a multi-configurable ``VGG-like'' model: we can generate multiple architectures depending on the depth $\delta$ (the larger, the deeper the model) and the number of convolutional filters per layer $2^{\gamma}$ (the larger, the wider the model). More details on the generated architectures can be found in Appendix.

\paragraph{Results} Fig.~\ref{fig:3D_Depth_50_Vanilla} shows the results at growing depth, with fixed $\gamma=5$. The sparse double descent becomes more and more evident for growing depths, while for shallow models we can only observe the traditional regime of sweet and collapsed phase. We observe a similar trend also in  Fig.~\ref{fig:3D_Width_50_Vanilla}, where the depth of the model is set to $\delta=1$ (two convolutional layers and one fully connected) while the width of the layer is increasing. With a fixed depth, increasing the width of the layers also reveals a double descent phenomenon: with low width, no SDD is observed, but it can be observed as $\gamma$ is increased. 

Typical analyses are performed, in the literature, in terms of the number of samples in the training set, or in terms of the width of the layers~\cite{nakkiran2021deep, chang2021provable, chen2021mitigating}. As the phenomenon also rises as a function of the model's depth, we believe SDD is more related to the number of parameters in a model rather than to the layer's organization and structure. This could motivate the occurrence of the sparse double descent, as the number of parameters is the varying quantity, while the width of the layer might not change.

\section{Distilling knowledge to avoid the sparse double descent}
\label{sec: KD avoid DD}
\begin{figure*}[t]
    \centering
    \begin{subfigure}{0.245\textwidth}
        \centering
        \includegraphics[width=\linewidth]{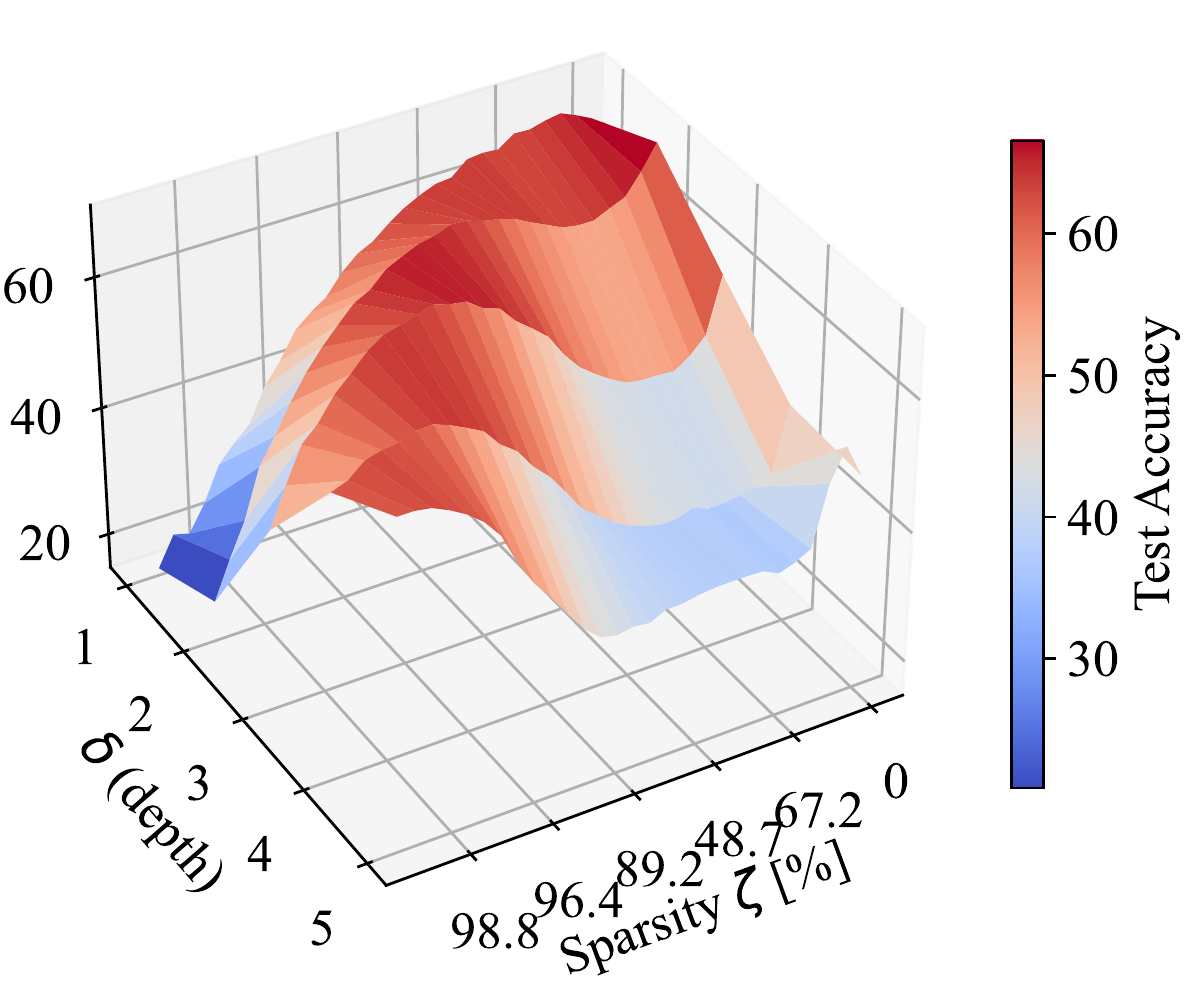}
        \caption{~}
        \label{fig:3D_Depth_50_Vanilla}
    \end{subfigure}
    \begin{subfigure}{0.245\textwidth}
        \centering
        \includegraphics[width=\linewidth]{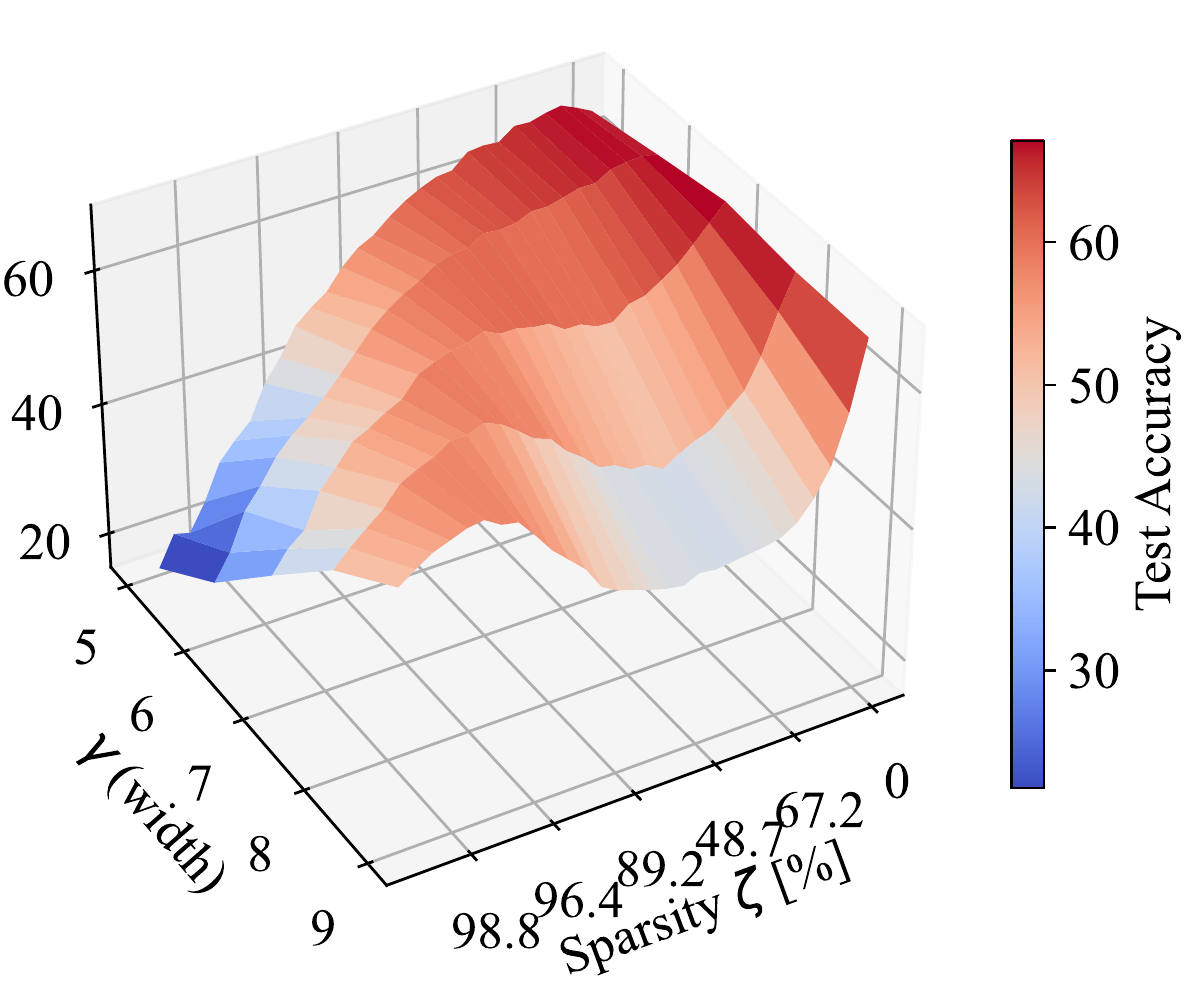}
        \caption{~}
        \label{fig:3D_Width_50_Vanilla}
    \end{subfigure}
    \begin{subfigure}{0.245\textwidth}
        \includegraphics[width=\linewidth]{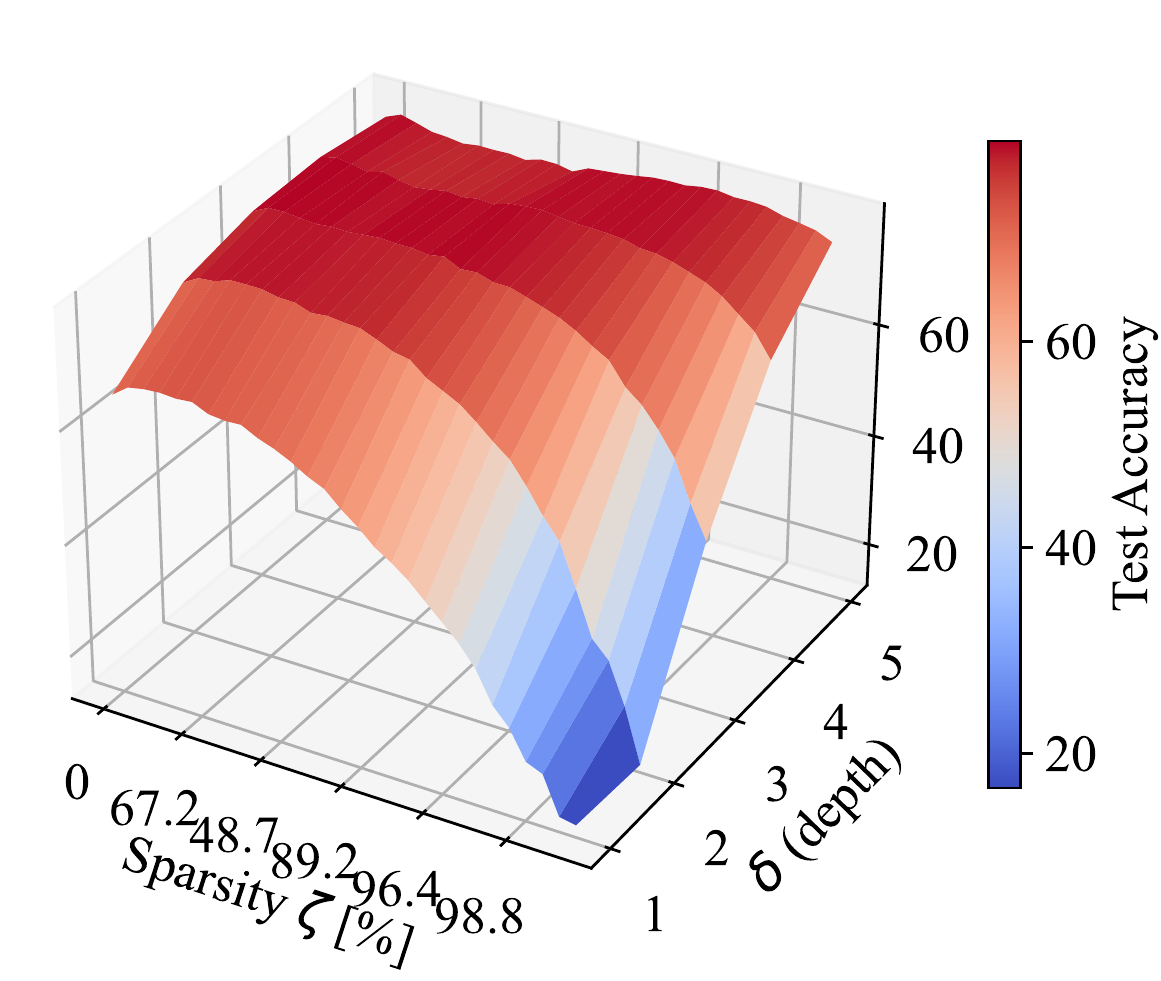}
        \caption{~}
        \label{fig:3D_Depth_50_KD}
    \end{subfigure}
    \begin{subfigure}{0.245\textwidth}
        \includegraphics[width=\linewidth]{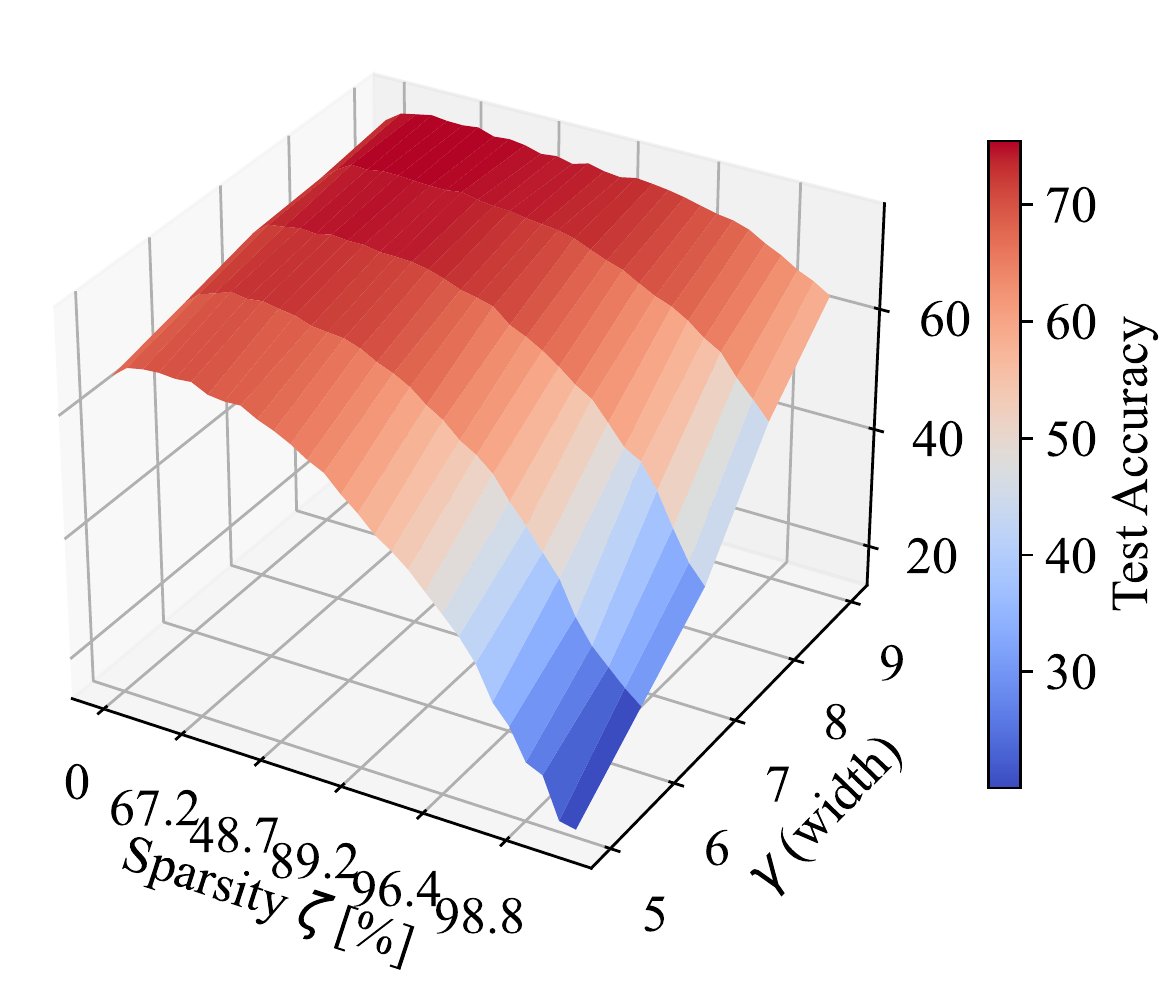}
        \caption{~}
        \label{fig:3D_Width_50_KD}
    \end{subfigure}
    \caption{Performance of VGG-like model, vanilla-trained (a, b), distilled from a sparse teacher (c, d), varying the depth $\delta$ (a, c) and the width $\gamma$ (b, d) on CIFAR-10 with $\varepsilon=50\%$.}
\end{figure*}

\begin{figure*}[t]
    \begin{subfigure}{0.33\textwidth}
        \includegraphics[width=0.95\textwidth]{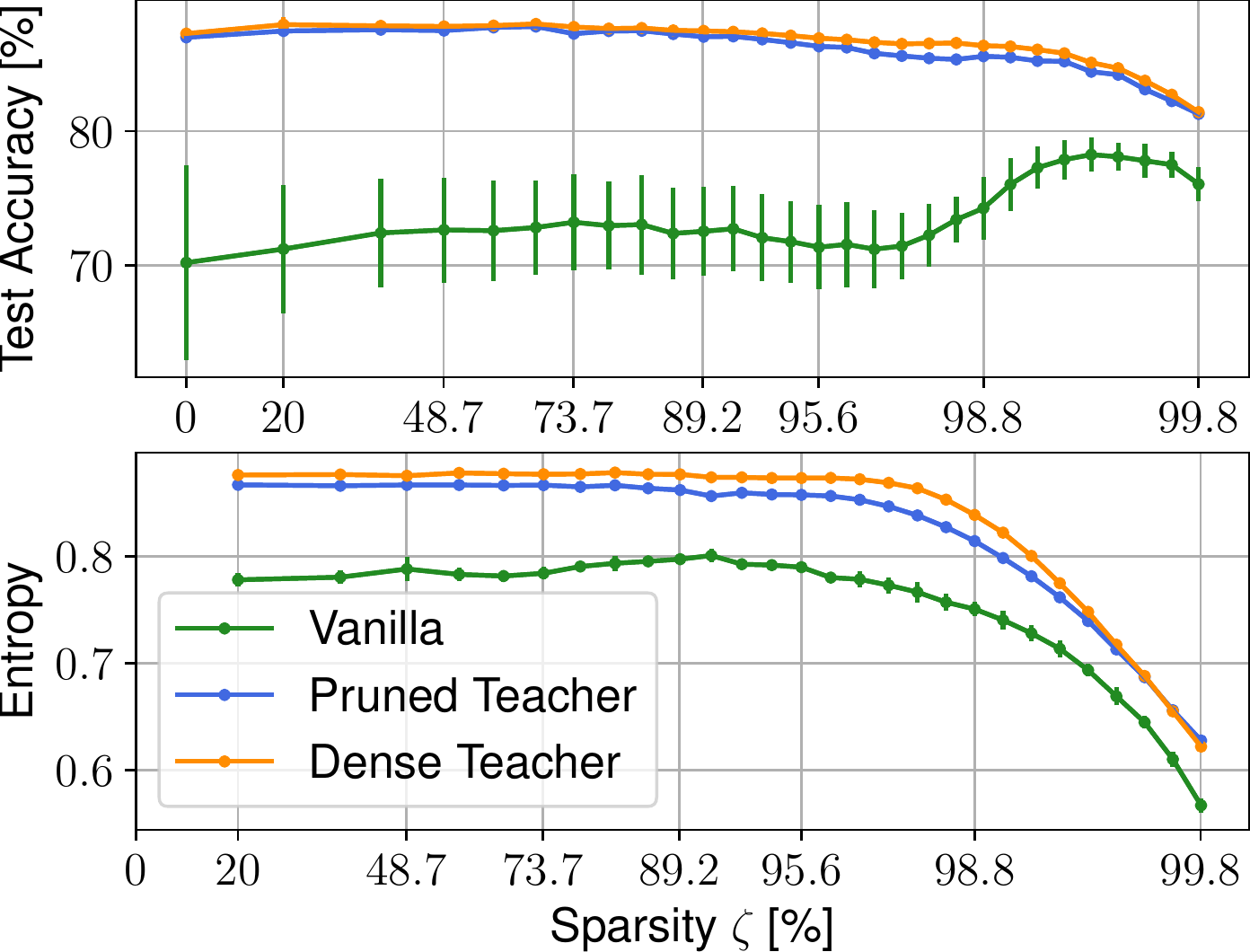}
        \caption{~}
        \label{fig:CIFAR-10_10_depth_5_width_32}
    \end{subfigure}
    \begin{subfigure}{0.33\textwidth}
        \includegraphics[width=0.95\textwidth]{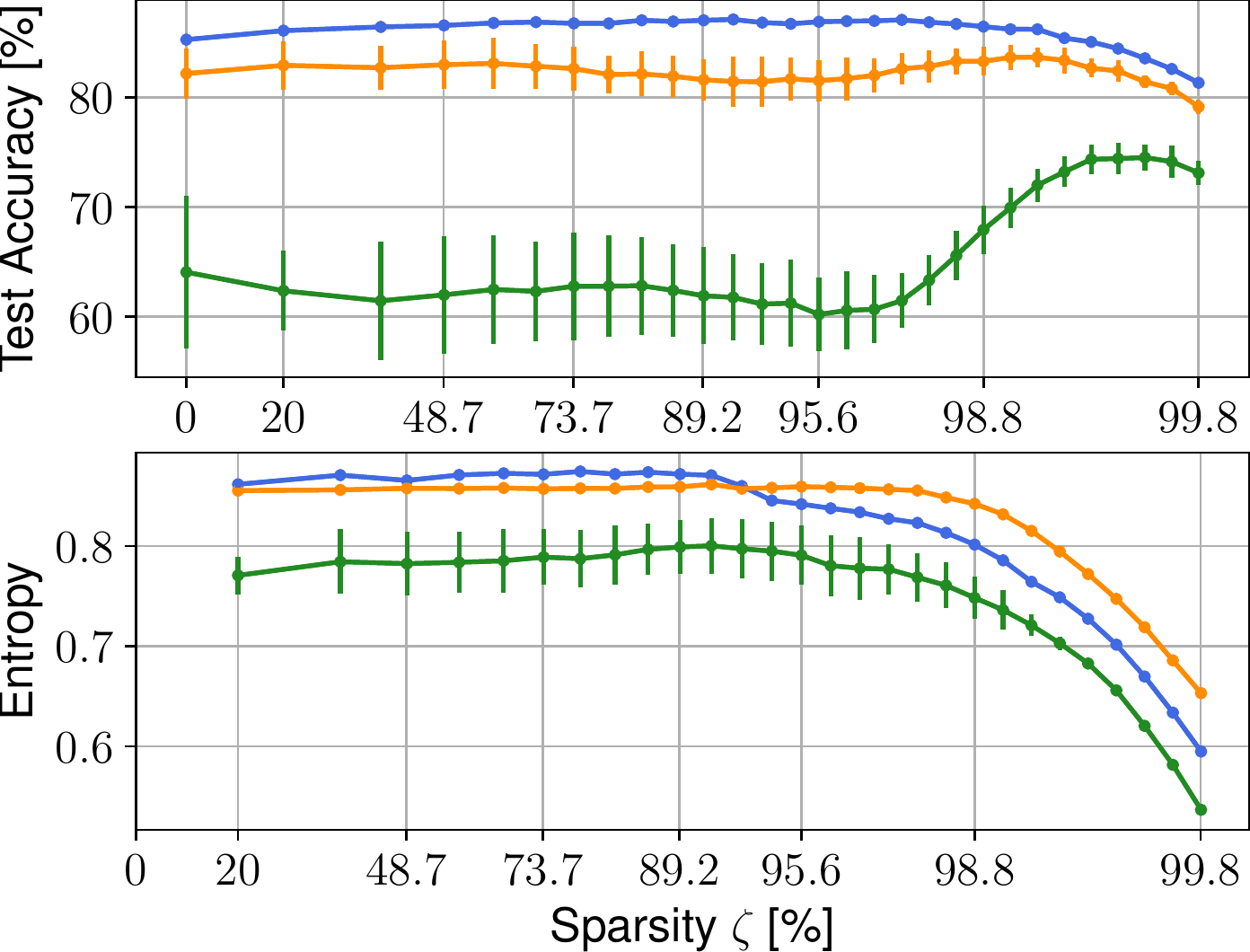}
        \caption{~}
        \label{fig:CIFAR-10_20_depth_5_width_32}
    \end{subfigure}
    \begin{subfigure}{0.33\textwidth}
        \includegraphics[width=0.95\textwidth]{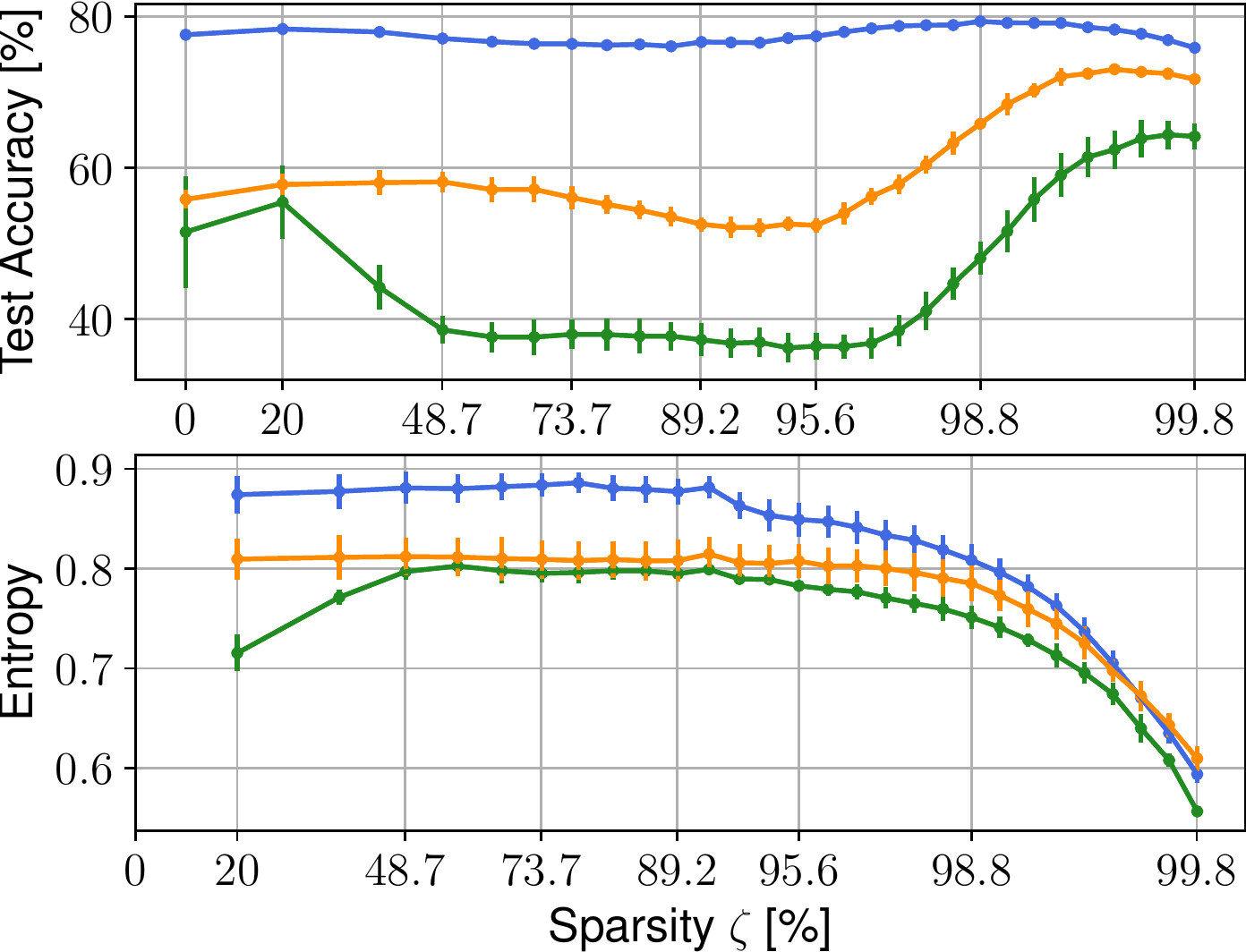}
        \caption{~}
        \label{fig:CIFAR-10_50_depth_5_width_32}
    \end{subfigure}
    \begin{subfigure}{0.33\textwidth}
        \includegraphics[width=0.95\textwidth]{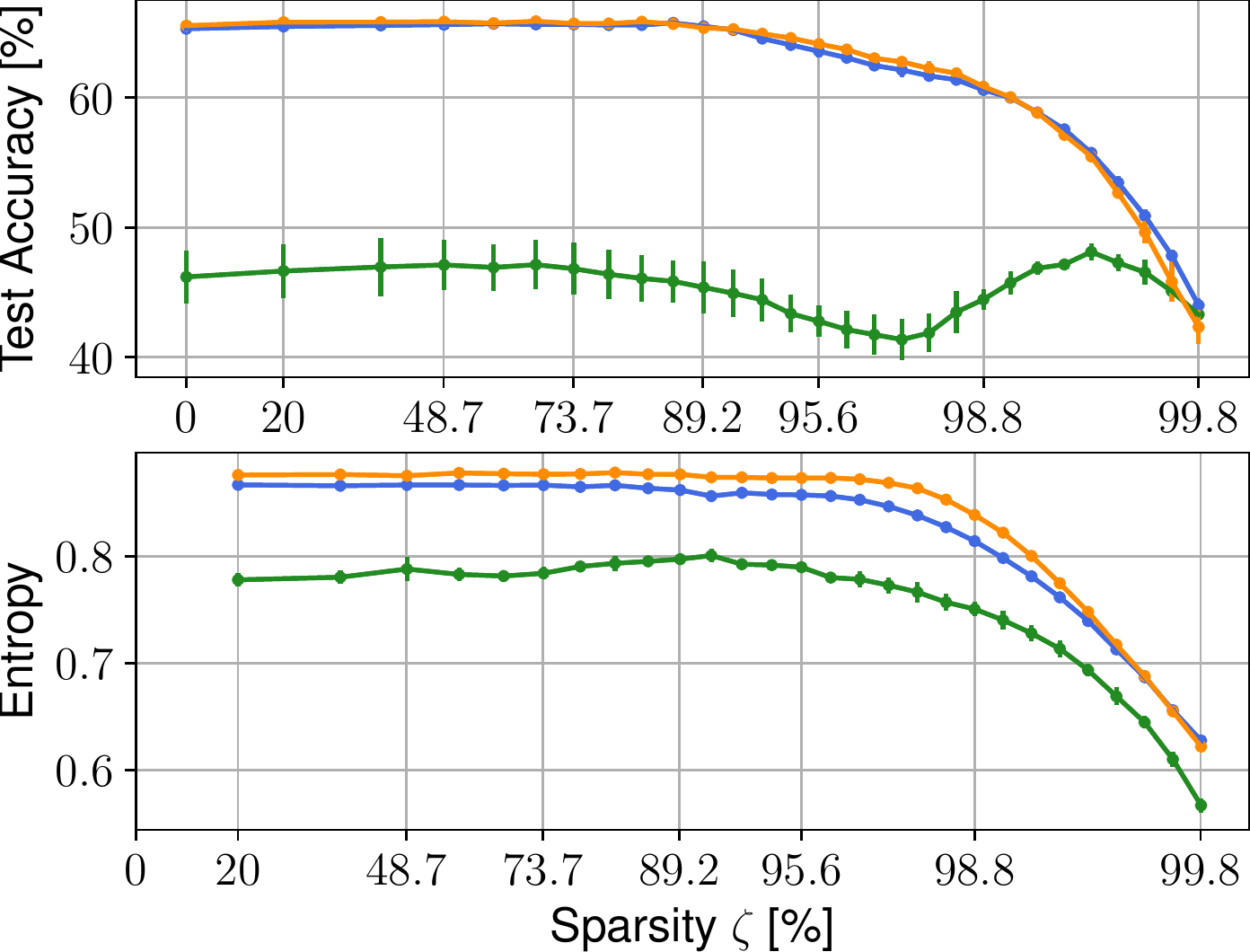}
        \caption{~}
        \label{fig:CIFAR-100_10_depth_5_width_32}
    \end{subfigure}
    \begin{subfigure}{0.33\textwidth}
        \includegraphics[width=0.95\textwidth]{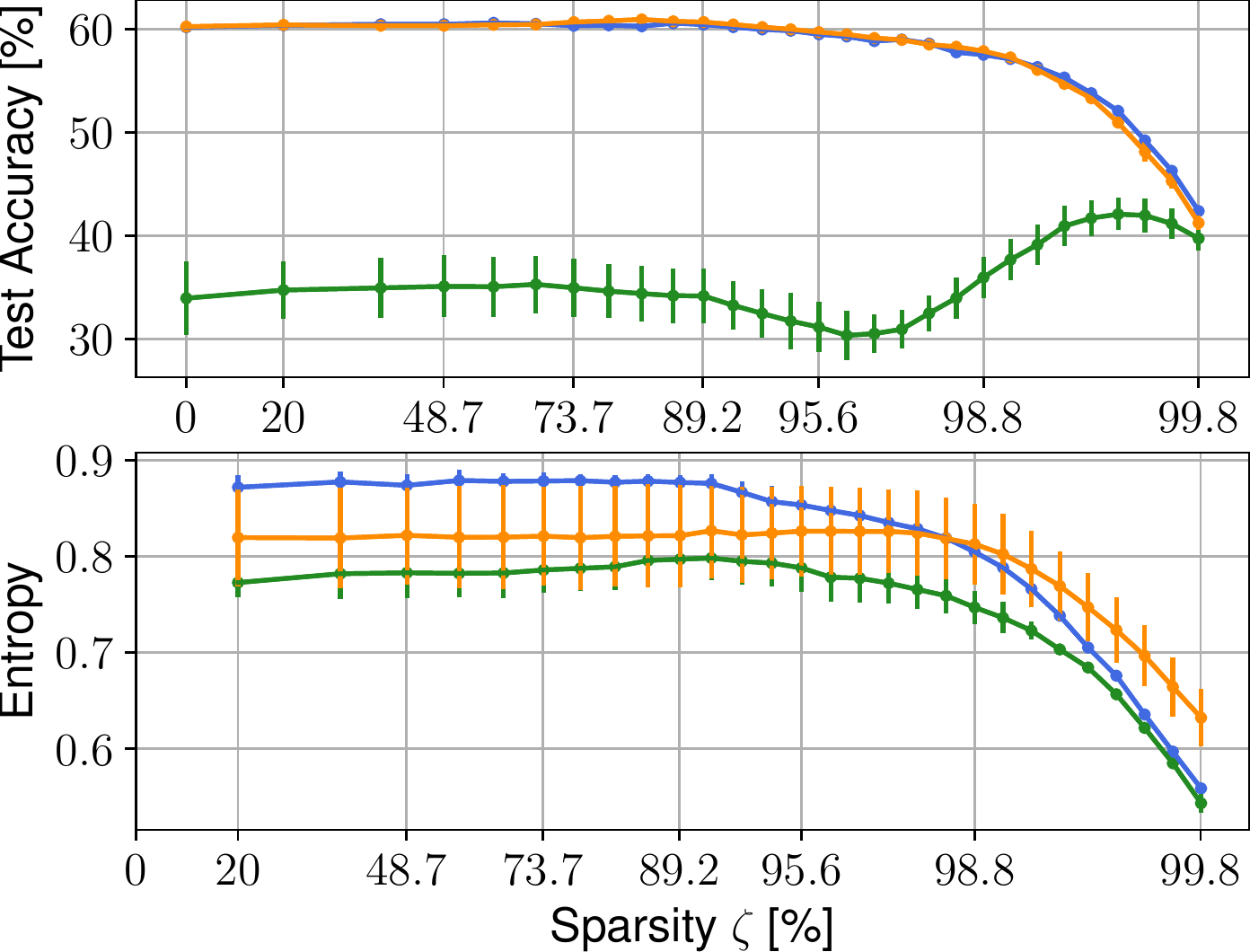}
        \caption{~}
        \label{fig:CIFAR-100_20_depth_5_width_32}
    \end{subfigure}
    \begin{subfigure}{0.33\textwidth}
        \includegraphics[width=0.95\textwidth]{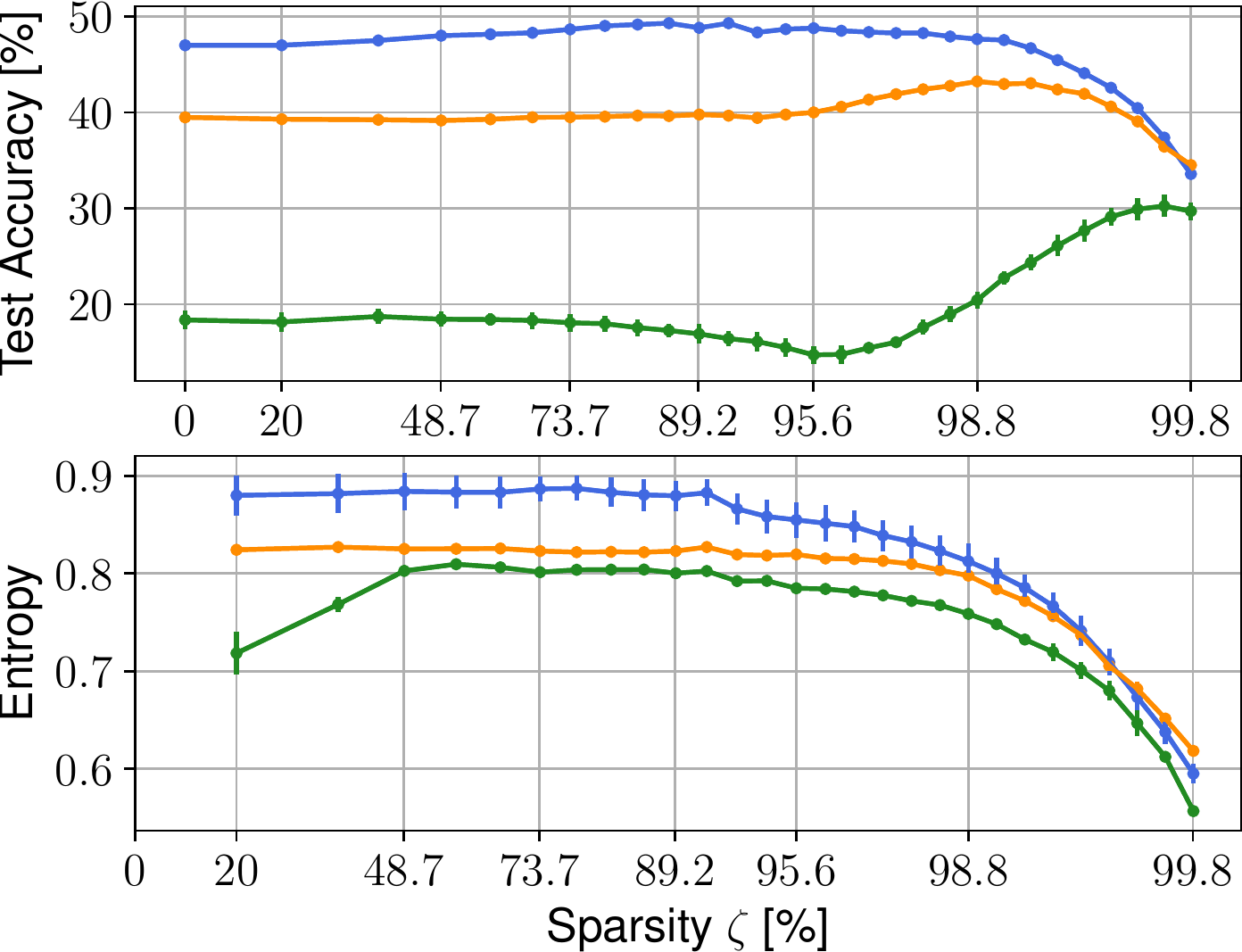}
        \caption{~}
        \label{fig:CIFAR-100_50_depth_5_width_32}
    \end{subfigure}
    \caption{Performance of the VGG-like model on CIFAR-10~(a, b, c) and CIFAR-100~(d, e, f) for different label noises. \textbf{Left:}~$~{\varepsilon=10\%}$. \textbf{Middle:} $\varepsilon=20\%$. \textbf{Right:} $\varepsilon=50\%$.}
    \label{fig:mainres}
\end{figure*}

Although a standard $\ell_2$ regularization approach can already positively contribute to dodging SDD,\footnote{We also present a study on other types of regularization strategies towards avoidance of SDD in Appendix.} 
it also presents some flaws.  \cite{nakkiran2021optimal} showed that, for certain linear regression models with isotropic data distribution, optimally-tuned $\ell_2$ regularization can achieve monotonic test performance as either the sample size or the model size is grown. However, \cite{10222624} highlighted that in some image classification setups, like ResNet-18 on CIFAR datasets, SDD is still noticeable even if $\ell_2$ regularization is used. Moreover, this regularization has the big drawback of sacrificing the performance/sparsity trade-off~\cite{quetu2023sparse}.
Thus, the need to design a custom regularization to prevent SDD, while maintaining good performance, becomes evident. In this section, we present a learning scheme in which a student model is regularized by distilling knowledge from a sparse teacher in its best validation accuracy region (or even with a dense teacher), observing that the student is dodging SDD.

\subsection{Background on knowledge distillation} Knowledge distillation (KD) transfers the knowledge from a (deep) teacher to a (shallow) student network. There are several knowledge types: response-based, feature-based, and relation-based. While response-based knowledge uses the output of the last layer of the teacher model as the knowledge~\cite{hinton2015distilling}, the output of intermediate layers, which in convolutional neural networks are the feature maps, is used to supervise the training of the student model for feature-based knowledge~\cite{romero2014fitnets}. Relation-based knowledge further explores the relationships between different layers or data samples~\cite{yim2017gift}. In this work, we will focus on the more flexible and easy-to-deploy response-based knowledge.

Distilling the knowledge from a teacher model, besides enhancing the student's performance, can also implicitly transmit good regularization properties. While \cite{yuan2020revisiting} observe that good students could also improve the teacher's performance, and \cite{furlanello2018born} observe improvement in distilling knowledge from students in a ``born again'' fashion, \cite{saglietti2022solvable} introduce a formal statistical physics framework that allows an analytic characterization of the properties of KD in shallow neural networks. By using Gaussian Mixture Models, it is shown that, without any fine-tuning at the level of the student loss function, KD allows a transfer of the (possibly fine-tuned) regularization properties of the teacher, even if the two models are mismatched, and even if the regularization strategy in the teacher training is not explicitly known. This work motivates us, in the quest for transmitting an implicit teacher's regularization (as being either in the well-generalizing light phase or the sweet phase for high label noise) to a student, in the attempt to escape from the SDD. Indeed, \cite{saglietti2022solvable} highlight a noticeable enhancement in the distillation test error when the transferred outputs are produced by an optimal teacher, with respect to a direct $\ell_2$ regularization: the gap with the optimal generalization bound is nearly closed. KD enables the student model to reach a performance unachievable with other regularization approaches, inheriting the teacher's regularization implicitly.

\subsection{Combining Knowledge Distillation and Network Pruning}
Several works have already explored the combination of KD and pruning. Towards generalization enhancement, in the context of working with reduced size of the train set, \cite{zhou2022efficient} propose a ``progressive feature distribution distillation'', which consists of first collecting a student network by pruning a trained network, and then distilling the feature distribution into the student.

A large slice of work combining the two techniques targets the final model's size reduction. For example, \cite{cui2021joint} combine structured pruning and dense knowledge distillation techniques to significantly compress an original large language model into a deep compressed shallow network, and \cite{wei22_interspeech} follow a similar strategy for speech enhancement. Still working on language models, \cite{kim2016sequence} reveal that standard KD applied to word-level prediction can be effective for Neural Machine Translation, and applied weight pruning on top of it to decrease the number of parameters.
Image processing-related works are, more applicative and oriented towards the enhancement of the performance on a sparsified model: \cite{chen2022knowledge} and \cite{aghli2021combining} are two works in this context. 
Leveraging on the beneficial, well-generalizing properties of both pruning and KD, \cite{park2022prune} provide several applicative examples where the ``prune, then distill'' pattern is practically very effective.

Motivated by \cite{saglietti2022solvable} and building on top of \cite{park2022prune} (although in a very different context), we formulate the following observation, which will drive our approach towards SDD dodging
\begin{obs}
    In an adversarial learning setup with knowledge distillation,  the teacher's response will act as a regularizer for the student model, which makes it avoid the sparse double descent. The student's performance, in a high-noise setup, is further enhanced when the knowledge is distilled from a sparse, well-performing teacher.
\end{obs}

\paragraph{Approach} In general, the objective function to be minimized while training a smaller student network in the KD framework is a linear combination of two losses, in the image classification setup: a standard cross-entropy loss $\mathcal{L}_\text{CE}$, which uses ``hard'' ground truth targets, and the Kullback-Leibler divergence loss $\mathcal{L}_\text{KL}$, calculated between the teacher’s and the student's predictions, eventually scaled by a temperature $\tau$:
\begin{equation}
    \label{eq:objKL}
    \mathcal{L}=(1-\alpha)\mathcal{L}_\text{CE}(\boldsymbol{y}^\text{s},\boldsymbol{\hat{y}})+\alpha \mathcal{L}_\text{KL}(\boldsymbol{y}^\text{s},\boldsymbol{y}^\text{t}, \tau),
\end{equation}
 where the apex ``$\text{s}$'' and ``$\text{t}$'' refer to the student and the teacher respectively, $\boldsymbol{\hat{y}}$ is the ground truth label, and $\alpha$ is the distillation hyper-parameter impacting on the weighted average between the distillation loss and the student loss. We employ the KD formulation loss as in~\cite{hinton2015distilling, kim2021comparing}.
 To distill the knowledge to some students, we require the teacher model to be already in its best-fit regime, achieved through Alg.~1 in Appendix, 
 and using the same sparsification process -but changing the objective function to \eqref{eq:objKL}-, we can train and sparsify the student model.

\paragraph{Results} For all the presented experiments here, $\alpha=0.8$ and $~{\tau=10}$. An ablation study on these two hyper-parameters and the set of all the other hyper-parameters used for the learning process are presented in Appendix. 
Fig.~\ref{fig:mainres} shows the results on CIFAR-10 and CIFAR-100 with $\varepsilon \in \left\{10\%, 20\%, 50\%\right\}$ using a VGG-like model with $2^\gamma=32$ and $\delta=5$ as student model, while ResNet-18 as teacher (whose training and performance is consistent to Fig.~\ref{fig:phases}). We observe, consistently for all the investigated noise rates, that the student model, trained with a vanilla training setup, always reveals the sparse double descent phenomenon. Nevertheless, the same student model, trained using a KD setup, can consistently avoid SDD. Furthermore, even if the knowledge is distilled from a dense teacher, SDD is always avoided, either in the case of best performance in the light or the sweet phase.
Observing the entropy for the model, it stays stationary and then decreases when the model enters the classical regime. Hence, once the entropy decreases, if the performance drops, the training can be stopped as the model enters the under-fitting regime.\footnote{We have performed experiments on three other datasets, without noise injection: CIFAR-100N, a human-annotated dataset, Flowers-102, a typically small dataset, and ImageNet in Appendix.}\\

\begin{table}[t]
    \centering
    \caption{Performance achieved and training computational cost of the student model with traditional approaches and our proposed scheme on CIFAR-10 with $\varepsilon=20\%$. The training cost of the teacher model is reported between parenthesis.}
    \label{tab:Computational Cost CIFAR-10 20}
    \resizebox{\linewidth}{!}{
    \begin{tabular}{c c c c c c }
    \toprule
    \bf Early & \bf \multirow{2}{*}{Distill} & \bf Distill from & \bf Training & \bf Test accuracy& \bf Sparsity \\
    \bf stop & & \bf pr. teach. & \bf [PFLOPs]($\boldsymbol{\downarrow}$) & \bf  [\%]($\boldsymbol{\uparrow}$) &\bf [\%] ($\boldsymbol{\uparrow}$)\\
    \midrule
    & & & 48.84 & 74.52 $\pm$ 1.20 & 99.62 \\
    $\checkmark$ & & & 4.88 & 60.23 $\pm$ 3.37 & 36.00 \\ 
    $\checkmark$ & $\checkmark$ & & 35.82 (+1.63) & 81.52 $\pm$ 1.85 & 99.26 \\
    $\checkmark$ & $\checkmark$ & $\checkmark$ & 35.82 (+47.21) & 86.89 $\pm$ 0.16 & 99.26 \\
    \bottomrule
    \end{tabular}
    }
\end{table}
\paragraph{Computation/performance trade-off} We compare traditional approaches with our proposed scheme on CIFAR-10 with $\varepsilon=20\%$ in Table~\ref{tab:Computational Cost CIFAR-10 20} in terms of performance achieved and training computational cost for the student model. 

In the vanilla case, early stop results in a model achieving low performance and with low sparsity. In order to extract a better performance, apparently, there is no other choice than pruning the model until all of its parameters are completely removed, which costs more than 48 PFLOPs. Our method achieves a model with high sparsity achieving more than 10\% improvement in performance with approximately 25\% computation less. We believe this is a core applicative contribution in real applications, where annotated datasets are small and noisy, and SDD can be easily observed. The same comparison for other setups can be found in Appendix.

\paragraph {Experiments with varying depth and width} In Fig.~\ref{fig:3D_Depth_50_KD} and Fig.~\ref{fig:3D_Width_50_KD} we report the distillation results on CIFAR-10 for $\varepsilon=50\%$, with a varying depth and width for the student (results for the other noise values are displayed in Appendix). While in Fig.~\ref{fig:3D_Depth_50_Vanilla} and Fig.~\ref{fig:3D_Width_50_Vanilla} we consistently observe that the vanilla-trained student model exhibits SDD as the width or depth is increased, we persistently notice in Fig.~\ref{fig:3D_Depth_50_KD} and Fig.~\ref{fig:3D_Width_50_KD}, however, that employing KD within the framework, the performance exhibits a monotonic behavior: the sparse double descent is dodged.

\paragraph{Limitations and future work} Leveraging a knowledge distillation scheme is a successful approach to transmitting good generalization properties from the teacher model to the student and enabling the dodge of SDD on the student itself. However, this method also presents some limits: in resource-constrained schemes, one might not afford to train a large teacher model. Nevertheless, our work is the first to enable the small model to enter the sweet phase instead of the critical phase, which cannot be achieved with traditional regularization strategies. Therefore, we let the exploration of more efficient approaches as future work.

\section{Conclusion}
\label{sec:Conclusion}
In this paper, we have studied and proposed a solution to dodge the sparse double descent. This phenomenon poses critical questions about where to find the model with the best performance in the low or the high sparsification regime. We observe an interesting correlation between the occurrence of the critical phase and the entropy of the activation's state for the training set - when not in the classical regime, the entropy is stationary. Although SDD challenges traditional early-stop schemes, the use of the network's entropy, which indicates the region we are navigating on, enables them back. We further observe that by leveraging a knowledge distillation scheme, not only the good generalization property of the teacher is transmitted to the student, but the student itself is no longer subject to SDD, due to the implicit regularization inherited by the teacher. 
We hope this work will ignite new research toward the improvement of learning strategies for deep, sparse models.

\section{Acknowledgements}
This work was performed using HPC resources from GENCI–IDRIS (Grant 2022-AD011013930).  Also, this research was partially funded by Hi!PARIS Center on Data Analytics and Artificial Intelligence.

\bibliography{main}

\onecolumn
\appendix

\section{Details on the architectures employed}
\label{Appendix: Details on the architectures employed}
To conduct extensive experiments to evaluate the impact of the width and the depth of the model on double descent, we have defined a model, based on a VGG-like architecture composed of blocks. Each block is composed of a 2D convolutional layer, followed by a ReLU activation function and a batch-normalization layer. 

Every VGG-like model is defined iteratively depending on the value of $\gamma$ and $\delta$.
$\delta$ stands for the number of (two blocks followed by a max-pooling layer). Hence, increasing $\delta$ by one is equivalent to adding two blocks ended by a max-pooling layer.
$\gamma$ is the power of two, setting the width of the convolutional layer. Thus, adding one to $\gamma$, is equivalent to multiplying the number of filters in the convolutional layer by 2.
The last two layers of the VGG-like model architecture are always an adaptative average pooling layer and a linear layer.
A summary of the architecture of the VGG-like model according to $\gamma$ and $\delta$ can be found in Table~\ref{tab:details on the simple model architecture}.

\begin{table}[H]
    \centering
    \begin{tabular}{|c|c|c|c|c|}
    \hline
    \multicolumn{5}{|c|}{\bf Possible depth configurations}\\
    \multicolumn{1}{|c}{$\delta=1$} & \multicolumn{1}{c}{$\delta=2$} & \multicolumn{1}{c}{$\delta=3$} & \multicolumn{1}{c}{$\delta=4$} & \multicolumn{1}{c|}{$\delta=5$} \\
    \hline
    \hline
    \multicolumn{5}{|c|}{Input (RGB image)}\\
    \hline
    Conv2d -$2^\gamma$ &  Conv2d -$2^\gamma$ & Conv2d -$2^\gamma$ & Conv2d -$2^\gamma$ & Conv2d -$2^\gamma$ \\
    ReLU &  ReLU & ReLU & ReLU & ReLU \\
    BatchNorm2d &  BatchNorm2d & BatchNorm2d & BatchNorm2d & BatchNorm2d \\
    Conv2d - $2^{\gamma}$ &  Conv2d - $2^{\gamma}$ & Conv2d - $2^{\gamma}$ & Conv2d - $2^{\gamma}$ & Conv2d - $2^{\gamma}$ \\
    ReLU &  ReLU & ReLU & ReLU & ReLU \\
    BatchNorm2d &  BatchNorm2d & BatchNorm2d & BatchNorm2d & BatchNorm2d \\
    \cline{2-5}
    \tikzmark{a}&\multicolumn{4}{c|}{Maxpool}\\
    \cline{2-5}
     &  Conv2d -$2^{\gamma+1}$ & Conv2d -$2^{\gamma+1}$ & Conv2d -$2^{\gamma+1}$ & Conv2d -$2^{\gamma+1}$ \\
    &  ReLU & ReLU & ReLU & ReLU \\
    &  BatchNorm2d & BatchNorm2d & BatchNorm2d & BatchNorm2d \\
    &  Conv2d - $2^{\gamma+1}$ & Conv2d - $2^{\gamma+1}$ & Conv2d - $2^{\gamma+1}$ & Conv2d - $2^{\gamma+1}$ \\
    &  ReLU & ReLU & ReLU & ReLU \\
    &  BatchNorm2d & BatchNorm2d & BatchNorm2d & BatchNorm2d \\
    \cline{3-5}
    &\tikzmark{c} &\multicolumn{3}{c|}{Maxpool}\\
    \cline{3-5}
    &  & Conv2d - $2^{\gamma+2}$ & Conv2d - $2^{\gamma+2}$ & Conv2d - $2^{\gamma+2}$ \\
    &  & ReLU & ReLU & ReLU \\
    &  & BatchNorm2d & BatchNorm2d & BatchNorm2d \\
    &  & Conv2d - $2^{\gamma+2}$ & Conv2d - $2^{\gamma+2}$ & Conv2d - $2^{\gamma+2}$ \\
    &  & ReLU & ReLU & ReLU \\
    &  & BatchNorm2d & BatchNorm2d & BatchNorm2d \\
    \cline{4-5}
    & & \tikzmark{e}&\multicolumn{2}{c|}{Maxpool}\\
    \cline{4-5}
    &  &  & Conv2d - $2^{\gamma+3}$ & Conv2d - $2^{\gamma+3}$ \\
    &  &  & ReLU & ReLU \\
    &  &  & BatchNorm2d & BatchNorm2d \\
    &  &  & Conv2d - $2^{\gamma+3}$ & Conv2d - $2^{\gamma+3}$ \\
    &  &  & ReLU & ReLU \\
    &  & & BatchNorm2d & BatchNorm2d \\
    \cline{5-5}
    & & & \tikzmark{g}&\multicolumn{1}{c|}{Maxpool}\\
    \cline{5-5}
    &  &  &  & Conv2d - $2^{\gamma+4}$ \\
    &  &  & & ReLU \\
    &  &  & & BatchNorm2d \\
    &  &  &  & Conv2d - $2^{\gamma+4}$ \\
    &  &  & & ReLU \\
    \tikzmark{b}& \tikzmark{d} & \tikzmark{f} & \tikzmark{h} & BatchNorm2d \\
    \hline
    \multicolumn{5}{|c|}{Adaptive Average Pooling}\\
    \hline
    \multicolumn{5}{|c|}{Fully Connected}\\
    \hline
    \end{tabular}
    \tikz[remember picture,overlay] \draw[->] (a.center -| b.center) -- (b.center);
    \tikz[remember picture,overlay] \draw[->] (c.center -| d.center) -- (d.center);
    \tikz[remember picture,overlay] \draw[->] (e.center -| f.center) -- (f.center);
    \tikz[remember picture,overlay] \draw[->] (g.center -| h.center) -- (h.center);
    \caption{The possible configurations used for the ``VGG-like model Architecture''.}
    \label{tab:details on the simple model architecture}
\end{table}

We include in Tab.~\ref{tab:Number of parameters} a summary of the model’s initial sizes used in the different experiments. Thanks to this initial size and the sparsity percentage of a model in a given experiment, one is able to have an idea of the total number of remaining/pruned weights.

\begin{table}[H]
    \centering
    \begin{tabular}{c c}
        \toprule
        \bf Architecture & \bf Number of parameters\\
        \midrule
        ResNet-18 & 11.1 M \\
        ResNet-50 & 23.5 M \\
        VGG-like ($\delta=1$, $\gamma=5$) & 26.0 K \\
        VGG-like ($\delta=1$, $\gamma=6$) & 70.3 K \\
        VGG-like ($\delta=1$, $\gamma=7$) & 214.4 K \\
        VGG-like ($\delta=1$, $\gamma=8$) & 723.7 K \\
        VGG-like ($\delta=1$, $\gamma=9$) & 2.6 M \\
        VGG-like ($\delta=2$, $\gamma=5$) & 97.3 K \\
        VGG-like ($\delta=3$, $\gamma=5$) & 350.6 K \\
        VGG-like ($\delta=4$, $\gamma=5$) & 1.3 M \\
        VGG-like ($\delta=5$, $\gamma=5$) & 4.9 M \\
        \bottomrule
    \end{tabular}
    \caption{Model's initial number of parameters for a given architecture.}
    \label{tab:Number of parameters}
\end{table}

\section{Details on the learning strategies employed}
The implementation details used in this paper are presented here. 
\label{Appendix: details on the learning strategies}
\subsection{Experimental details}

Like in \cite{SparseDoubleDescent} set up, for the ResNet-18 network, a modified version of the \texttt{torchvision} model is used: the first convolutional layer is set with a filter of size 3 × 3 and the max-pooling layer that follows has been eliminated to adapt ResNet-18 for CIFAR-10 and CIFAR-100. CIFAR-10 and CIFAR-100 are augmented with per-channel normalization, random horizontal flipping, and random shifting by up to four pixels in any direction. ImageNet is augmented with per-channel normalization, random horizontal flipping, random cropping, and resizing to 224.

In pruning experiments, all weights from convolutional and linear layers are set as prunable no matter the model architecture. Neither biases nor batch normalization parameters are pruned. 

The training hyperparameters used in the experiments are presented in Table \ref{tab: Learning strategies}. Our code can be found at \url{https://github.com/VGCQ/DSD2}.
\begin{table}[H]
\resizebox{\textwidth}{!}{
    \centering
    \begin{tabular}{c c c c c c c c c c c}
        \toprule
        \bf Model & \bf Dataset & \bf Epochs & \bf Batch & \bf Opt. & \bf Mom. & \bf LR & \bf Milestones & \bf Drop Factor & \bf Weight Decay & \bf Rewind Iter. \\
        \midrule
         ResNet-18 & CIFAR-10 & 160 & 128 & SGD & 0.9 & 0.1 & [80, 120] & 0.1 & 1e-4 & 1000 \\
         VGG-like & CIFAR-10 & 160 & 128 & SGD & 0.9 & 0.001 & [80, 120] & 0.1 & 1e-4 & 0
         \\
         ResNet-18 & CIFAR-100 & 160 & 128 & SGD & 0.9 & 0.1 & [80, 120] & 0.1 & 1e-4 & 1000
         \\
         VGG-like & CIFAR-100 & 160 & 128 & SGD & 0.9 & 0.001 & [80, 120] & 0.1 & 1e-4 & 0
         \\
         ResNet-18 & CIFAR-100N & 160 & 128 & SGD & 0.9 & 0.1 & [80, 120] & 0.1 & 1e-4 & 1000
         \\
         VGG-like & CIFAR-100N & 160 & 128 & SGD & 0.9 & 0.001 & [80, 120] & 0.1 & 1e-4 & 0
         \\
         ResNet-18 & Flowers102 & 160 & 64 & SGD & 0.9 & 0.001 & [80, 120] & 0.1 & 1e-5 & 0 \\
         ResNet-50 & ImageNet & 90 & 1024 & SGD & 0.9 & 0.1 & [30, 60] & 0.1 & 1e-4 & 0
         \\
         ResNet-18 & ImageNet & 90 & 1024 & SGD & 0.9 & 0.1 & [30, 60] & 0.1 & 1e-4 & 0
         \\
          \bottomrule
    \end{tabular}
    }
    \caption{Table of the different employed learning strategies.}
    \label{tab: Learning strategies}
\end{table}

\subsection{Algorithm}
First, we introduce the function PPTE in Alg.~\ref{Algo}, which is used in Alg.~\ref{Algo} and Alg.~\ref{Algo_early_stop}.  

The function first \textbf{P}runes the model $\mathcal{M}$(line~\ref{line:prune}) using some pruning strategy (we employ magnitude pruning, following~\cite{SparseDoubleDescent}), \textbf{P}erturb (line \ref{line:perturb} - weights can be rewound to initialization, randomly re-initialized, or not perturbed at all), and re-\textbf{T}rained on the training set $\Xi_{train}$ (line~\ref{line:retrain}). Then, it \textbf{E}valuates the performance of the model on the validation set $\Xi_{val}$ (line~\ref{line:evaluate}).
Finally, the function returns the model, represented by its weights $\boldsymbol{w}^\mathcal{M}$, as well as its performance on the validation set (line~\ref{line:return}).

\newpage
Moreover, we present our iterative pruning algorithm employed to reduce the dimensionality of the trained $\mathcal{M}$ in Alg.~\ref{Algo}.
\begin{algorithm}
\caption{Iterative pruning algorithm \& PPTE function.}
\label{Algo}
\begin{algorithmic}[1]
    \FUNCTION{Iterative pruning($\boldsymbol{w}^\mathcal{M}$, $\Xi$, $\zeta$, $\zeta_{\text{wall}}$)}{}
    \STATE $\boldsymbol{w}^\mathcal{M}\gets $Train($\boldsymbol{w}^\mathcal{M}$, $\Xi_{train}$) \alglinelabel{line:1}
    \STATE $\boldsymbol{w}_{\text{best}}^\mathcal{M}$ $\gets$ $\boldsymbol{w}^\mathcal{M}$
    \STATE best\_acc $\gets$ Evaluate($\zeta_{\text{current}}$, $\Xi_{val}$)
    \STATE $\zeta_{\text{current}}\gets \zeta$
    \WHILE{$\zeta_{\text{current}} < \zeta_{\text{wall}}$} \alglinelabel{line:6}
        \STATE $\boldsymbol{w}^\mathcal{M}$, this\_acc $\gets$ PPTE($\boldsymbol{w}^\mathcal{M}$, $\zeta_{\text{current}}$, $\Xi_{train}$, $\Xi_{val}$)\alglinelabel{line:function}
        \IF{this\_acc$>$best\_acc}
            \STATE $\boldsymbol{w}_{\text{best}}^\mathcal{M} \gets \boldsymbol{w}^\mathcal{M}$ \alglinelabel{line:12}
            \STATE best\_acc $\gets$ this\_acc
        \ENDIF
        \STATE $\zeta_{\text{current}} \gets 1 - (1-\zeta_{\text{current}}) (1-\zeta)$
    \ENDWHILE
    \STATE \textbf{return} $\boldsymbol{w}_{\text{best}}^\mathcal{M}$ \alglinelabel{line:17}
    \ENDFUNCTION
    \STATE
    \FUNCTION{PPTE($\boldsymbol{w}^\mathcal{M}$, $\zeta$, $\Xi_{train}$, $\Xi_{val}$)}{}
        \STATE $\boldsymbol{w}^\mathcal{M} \gets$ \textbf{P}rune($\boldsymbol{w}^\mathcal{M}$, $\zeta$) \alglinelabel{line:prune}
        \STATE $\boldsymbol{w}^\mathcal{M} \gets$ \textbf{P}erturb($\boldsymbol{w}^\mathcal{M}$) \alglinelabel{line:perturb}
        \STATE $\boldsymbol{w}^\mathcal{M}\gets $\textbf{T}rain($\boldsymbol{w}^\mathcal{M}$, $\Xi_{train}$)\alglinelabel{line:retrain}
        \STATE this\_acc $\gets$ \textbf{E}valuate($\zeta_{\text{current}}$, $\Xi_{val}$)\alglinelabel{line:evaluate}
    \STATE \textbf{return} $\boldsymbol{w}^\mathcal{M}$, this\_acc \alglinelabel{line:return}
    \ENDFUNCTION
\end{algorithmic}
\end{algorithm}

The first step is to train the dense model (line~\ref{line:1}). Until it has reached the desired sparsity percentage $\zeta_{\text{wall}}$ (line~\ref{line:6}), the model is iteratively pruned, perturbed, re-trained on $\Xi_{train}$ and evaluated on the validation set $\Xi_{val}$ (line~\ref{line:function}). When $\zeta_{\text{wall}}$ is reached, the algorithm returns the model parameters $\boldsymbol{w_{\text{best}}}$, which achieve the best performance on the validation set $\Xi_\text{val}$ (line~\ref{line:17}).
\newpage
\section{Ablation study for $\alpha$ and $\tau$}
\label{sec:ablation}
In order to choose the set of parameters $\alpha$ and $\tau$ which are used in our learning framework, we carried out an ablation study over these 2 parameters.
We report in Tab.~\ref{tab:Ablation_alpha_tau} the results of the validation accuracy achieved by the simple model on CIFAR-10 with $\varepsilon=10\%$, $\varepsilon=20\%$ and $\varepsilon=50\%$ for different sets of $\alpha$ and $\tau$.
To choose the final $\alpha$ and $\tau$, we compute the average validation accuracy obtained for the 3 noise ratios. We observe that the best validation accuracy is achieved for $\alpha=0.8$ and $\tau=10$.

\begin{table}[H]
    \centering
    \begin{tabular}{c c c c}
        \toprule
        \bf Noise rate $\boldsymbol{\varepsilon}$ & $\boldsymbol{\alpha}$ & \bf Temperature $\boldsymbol{\tau}$ & \bf Validation Accuracy\\
        \midrule
         \multirow{8}{*}{10\%} & 0.5 &  & 86.77 \\
         & 0.7 & 10 & 86.06 \\
         & 0.8 &  & \textbf{87.33} \\
         & 0.9 &  & 86.69 \\ \cline{2-4}
         & 0.5 &  & 86.26 \\ 
         & 0.7 & 20 & 86.33 \\
         & 0.8 &  & 86.76 \\
         & 0.9 &  & 86.92 \\
         \midrule
         \multirow{8}{*}{20\%}& 0.5 & & 84.64 \\
         & 0.7 & 10 & \textbf{85.36} \\
         & 0.8 & & 85.24 \\
         & 0.9 & & 84.66 \\ \cline{2-4}
         & 0.5 & & 84.51 \\
         & 0.7 & 20 & 84.90 \\
         & 0.8 & & 85.27 \\
         & 0.9 & & 85.28 \\
         \midrule
         \multirow{8}{*}{50\%}& 0.5 & & 77.72 \\
         & 0.7 & 10 & \textbf{79.23} \\
         & 0.8 & & 79.09 \\
         & 0.9 & & 78.91 \\ \cline{2-4}
         & 0.5 & & 76.46 \\ 
         & 0.7 & 20 & 77.53 \\
         & 0.8 & & 78.67 \\
         & 0.9 & & 78.98 \\
         \midrule
         & 0.5 & & 83.04\\
         & 0.7 & 10 & 83.55\\
         & 0.8 & & \textbf{83.89}\\
         Average on $\varepsilon$& 0.9 & & 83.42\\ \cline{2-4}
         values & 0.5 & & 82.41\\
         & 0.7 & 20 & 82.92\\
         & 0.8 & & 83.57\\
         & 0.9 & & 83.73\\
         \bottomrule
    \end{tabular}
\caption{Ablation study over $\alpha$ and $\tau$ on CIFAR-10 for the VGG-like architecture. The best performance over the 3 noise rates is achieved for $\tau=10$ and $\alpha=0.8$.}
    \label{tab:Ablation_alpha_tau}
\end{table}
\section{Comparison with other regularization approaches towards avoidance of SDD}

\begin{figure}[h!]
    \centering
    \includegraphics[width=0.80\linewidth]{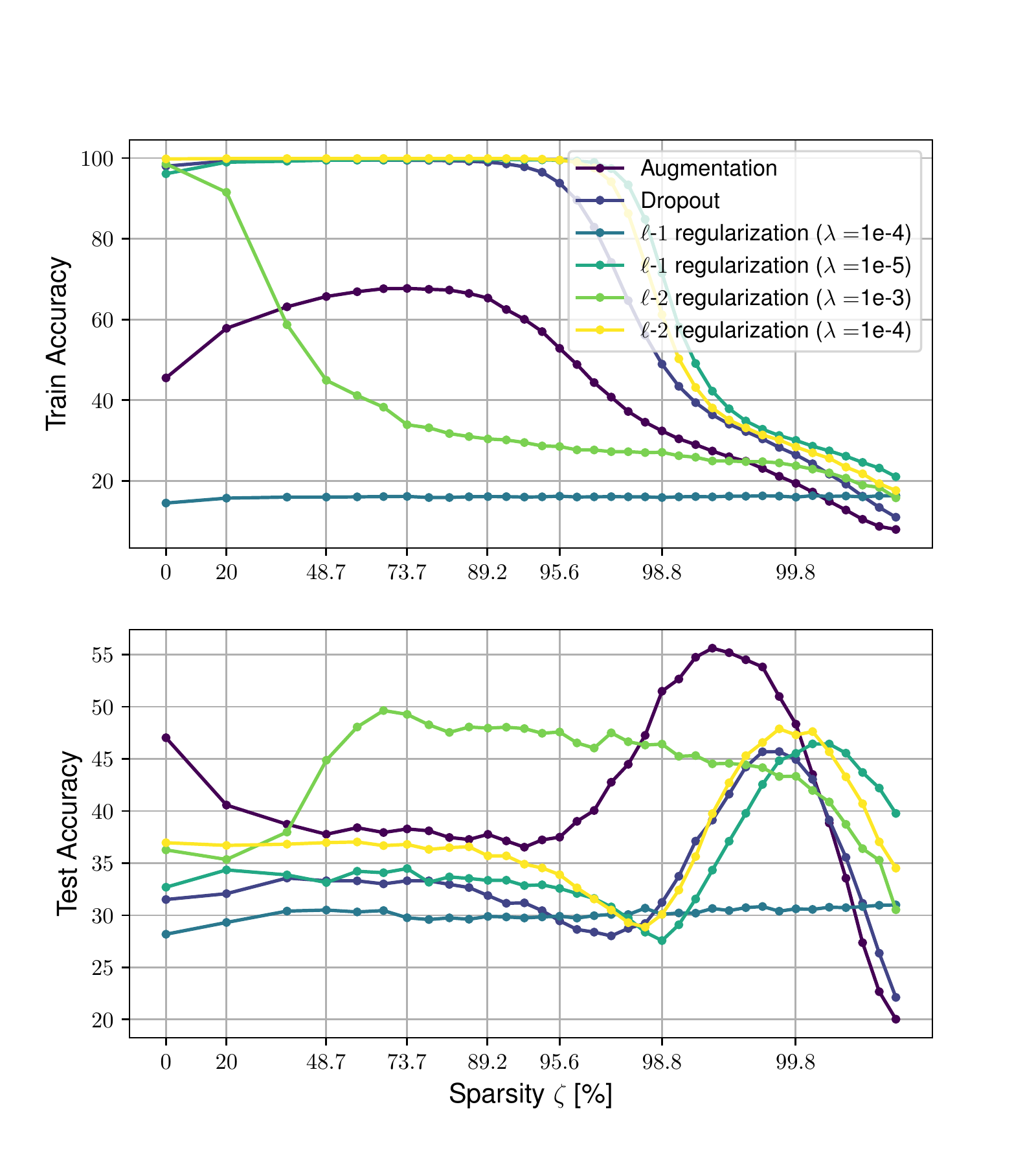}
    \caption{Comparison with other regularization approaches for a ResNet-18 on CIFAR-100 with $\varepsilon=50\%$.}
    \label{fig:other_reg}
\end{figure}

\cite{nakkiran2021optimal} showed in regression tasks that an optimal $\ell_2$ regularization can help mitigate DD. However, a recent work \cite{10222624} highlights that in image classification, the problem is not easily overcome and this regularization is not sufficient enough to completely lessen the phenomenon: a more complex approach has to be leveraged.
To compare our framework with other existing regularization methods, we conduct here a comparison with other regularization approaches: $\ell_1$, $\ell_2$, dropout, and data augmentation. The results for a ResNet-18 are presented in Fig.\ref{fig:other_reg} on CIFAR-100 with $\varepsilon = 50\%$. 

The use of any of these regularizations isn’t helping to relieve the phenomenon. Indeed, SDD occurs in the experiments using data augmentation (the training dataset is augmented with AutoAugment and CutOut). Moreover, even if a dropout layer before the classifier with a probability of 0.5 is used, SDD still appears. Finally, for $\ell_1$ and $\ell_2$ regularizations weighted by $\lambda$, we show that for $\lambda=1e-5$ and $\lambda=1e-4$  respectively, SDD is not mitigated. Moreover, by increasing their value to respectively 1e-4 and 1e-3, the training fails in the case of $\ell_1$, and the phenomenon is just shifted to the left (confirming the results shown in \cite{10222624}).
\newpage
\section{Details on the computation resources and the compression/performance tradeoff.}
\label{Appendix:computation_cost}

A model is subject to SDD if its over-parametrization is massive~\cite{SparseDoubleDescent}: it is impossible to effectively stop the learning/compression with any early-stop criteria, as the accuracy curve will invert, at some point, its trend. Guaranteeing the SSD is dodged will avoid such non-monotonic behavior, enabling back early stop approaches. To illustrate the motivation, let us compare traditional approaches with our proposed scheme presented in Alg.~\ref{Algo_early_stop} on CIFAR-10 and CIFAR-100 with $\varepsilon \in \left\{10\%, 20\%, 50\%\right\}$ here below. 

\begin{table}[!h]
    \centering
    \resizebox{\textwidth}{!}{
    \begin{tabular}{c c c c c c }
    \toprule
    \bf \multirow{2}{*}{Early stop} & \bf \multirow{2}{*}{Distillation} & \bf Distillation from & \bf Training FLOPs & \bf Test accuracy& \bf \multirow{2}{*}{Sparsity [\%] ($\boldsymbol{\uparrow}$)} \\
    & & \bf pruned teacher & \bf [PFLOPs]($\boldsymbol{\downarrow}$) & \bf  [\%]($\boldsymbol{\uparrow}$) & \\
    \midrule
    & & & 48.84 & 76.04 $\pm$ 1.30 & 99.52\\
    $\checkmark$ & & & 48.84 & 76.04 $\pm$ 1.30 & 99.80\\ 
    $\checkmark$ & $\checkmark$ & & 48.84 (+1.63) & 81.42 $\pm$ 0.15 & 99.80 \\
    $\checkmark$ & $\checkmark$ & $\checkmark$ & 48.84 (+9.77) & 81.29 $\pm$ 0.23 & 99.80 \\
    \bottomrule
    \end{tabular}
    }
    \caption{Performance achieved and training computational cost of traditional approaches and our proposed scheme on CIFAR-10 with $\varepsilon=10\%$.}
\end{table}

\begin{table}[!h]
    \centering
    \resizebox{\textwidth}{!}{
    \begin{tabular}{c c c c c c }
    \toprule
    \bf \multirow{2}{*}{Early stop} & \bf \multirow{2}{*}{Distillation} & \bf Distillation from & \bf Training FLOPs & \bf Test accuracy& \bf \multirow{2}{*}{Sparsity [\%] ($\boldsymbol{\uparrow}$)} \\
    & & \bf pruned teacher & \bf [PFLOPs]($\boldsymbol{\downarrow}$) & \bf  [\%]($\boldsymbol{\uparrow}$) & \\
    \midrule
    & & & 48.84 & 74.52 $\pm$ 1.20 & 99.62 \\
    $\checkmark$ & & & 4.88 & 60.23 $\pm$ 3.37 & 36.00 \\ 
    $\checkmark$ & $\checkmark$ & & 35.82 (+1.63) & 81.52 $\pm$ 1.85 & 99.26 \\
    $\checkmark$ & $\checkmark$ & $\checkmark$ & 35.82 (+47.21) & 86.89 $\pm$ 0.16 & 99.26 \\
    \bottomrule
    \end{tabular}
    }
    \caption{Performance achieved and training computational cost of traditional approaches and our proposed scheme on CIFAR-10 with $\varepsilon=20\%$.}
\end{table}

\begin{table}[!h]
    \centering
    \resizebox{\textwidth}{!}{
    \begin{tabular}{c c c c c c }
    \toprule
    \bf \multirow{2}{*}{Early stop} & \bf \multirow{2}{*}{Distillation} & \bf Distillation from & \bf Training FLOPs & \bf Test accuracy& \bf \multirow{2}{*}{Sparsity [\%] ($\boldsymbol{\uparrow}$)} \\
    & & \bf pruned teacher & \bf [PFLOPs]($\boldsymbol{\downarrow}$) & \bf  [\%]($\boldsymbol{\uparrow}$) & \\
    \midrule
    & & & 48.84 & 64.37 $\pm$ 1.89  & 99.76\\
    $\checkmark$ & & & 6.51 & 38.57 $\pm$ 1.85 & 48.80\\ 
    $\checkmark$ & $\checkmark$ & & 48.84 (+1.63) & 71.72 $\pm$ 0.36 & 99.80\\
    $\checkmark$ & $\checkmark$ & $\checkmark$ & 48.84 (+52.09) & 75.87 $\pm$ 0.18 & 99.80\\
    \bottomrule
    \end{tabular}
    }
    \caption{Performance achieved and training computational cost of traditional approaches and our proposed scheme on CIFAR-10 with $\varepsilon=50\%$.}
\end{table}


\begin{table}[!h]
    \centering
    \resizebox{\textwidth}{!}{
    \begin{tabular}{c c c c c c }
    \toprule
    \bf \multirow{2}{*}{Early stop} & \bf \multirow{2}{*}{Distillation} & \bf Distillation from & \bf Training FLOPs & \bf Test accuracy& \bf \multirow{2}{*}{Sparsity [\%] ($\boldsymbol{\uparrow}$)} \\
    & & \bf pruned teacher & \bf [PFLOPs]($\boldsymbol{\downarrow}$) & \bf  [\%]($\boldsymbol{\uparrow}$) & \\
    \midrule
    & & & 48.84 & 48.13 $\pm$ 0.68 & 99.53\\
    $\checkmark$ & & & 19.54 & 44.93 $\pm$ 1.85 &  91.41\\ 
    $\checkmark$ & $\checkmark$ & & 22.79 (+1.63) & 64.60 $\pm$ 0.20 & 94.50\\
    $\checkmark$ & $\checkmark$ & $\checkmark$ &  22.79 (+6.51) & 64.07 $\pm$ 0.33 & 94.50\\
    \bottomrule
    \end{tabular}
    }
    \caption{Performance achieved and training computational cost of traditional approaches and our proposed scheme on CIFAR-100 with $\varepsilon=10\%$.}
\end{table}

\newpage
\begin{table}[!h]
    \centering
    \resizebox{\textwidth}{!}{
    \begin{tabular}{c c c c c c }
    \toprule
    \bf \multirow{2}{*}{Early stop} & \bf \multirow{2}{*}{Distillation} & \bf Distillation from & \bf Training FLOPs & \bf Test accuracy& \bf \multirow{2}{*}{Sparsity [\%] ($\boldsymbol{\uparrow}$)} \\
    & & \bf pruned teacher & \bf [PFLOPs]($\boldsymbol{\downarrow}$) & \bf  [\%]($\boldsymbol{\uparrow}$) & \\
    \midrule
    & & & 48.84 & 52.06 $\pm$ 1.57 & 99.62\\
    $\checkmark$ & & & 24.42 & 30.33 $\pm$ 2.39 & 96.48\\ 
    $\checkmark$ & $\checkmark$ & & 30.93 (+1.63) & 58.26 $\pm$ 0.45 & 98.56\\
    $\checkmark$ & $\checkmark$ & $\checkmark$ & 30.93 (+14.65) & 57.73 $\pm$ 0.29 & 98.56\\
    \bottomrule
    \end{tabular}
    }
    \caption{Performance achieved and training computational cost of traditional approaches and our proposed scheme on CIFAR-100 with $\varepsilon=20\%$.}
\end{table}

\begin{table}[!h]
    \centering
    \resizebox{\textwidth}{!}{
    \begin{tabular}{c c c c c c }
    \toprule
    \bf \multirow{2}{*}{Early stop} & \bf \multirow{2}{*}{Distillation} & \bf Distillation from & \bf Training FLOPs & \bf Test accuracy& \bf \multirow{2}{*}{Sparsity [\%] ($\boldsymbol{\uparrow}$)} \\
    & & \bf pruned teacher & \bf [PFLOPs]($\boldsymbol{\downarrow}$) & \bf  [\%]($\boldsymbol{\uparrow}$) & \\
    \midrule
    & & & 48.84 & 30.24 $\pm$ 1.16 & 99.75\\
    $\checkmark$ & & & 22.79 & 14.72 $\pm$ 0.93  & 95.60\\ 
    $\checkmark$ & $\checkmark$ & & 42.33 (+1.63) & 39.06 $\pm$ 0.19 & 99.70\\
    $\checkmark$ & $\checkmark$ & $\checkmark$ & 42.33 (+45.58) & 40.45 $\pm$ 0.21 & 99.70\\
    \bottomrule
    \end{tabular}
    }
    \caption{Performance achieved and training computational cost of traditional approaches and our proposed scheme on CIFAR-100 with $\varepsilon=50\%$.}
\end{table}

In the vanilla case (second lines), early stop results in a model achieving low performance and low sparsity. In order to extract a better performance, apparently, there is no other choice than pruning the model until all of its parameters are completely removed, which results in a \textit{big computation overhead} (first lines).

Our method achieves a model with high sparsity resulting in better performance with less computation compared to the vanilla case. We believe this is a core applicative contribution in real applications, where annotated datasets are small and noisy, and SDD can be easily observed.

\newpage
\section{Correlation between $\boldsymbol{w}_{best}$ and $\varepsilon$.}
\label{Appendix:correlation_noise}

We include below a deeper analysis of the correlation between $\boldsymbol{w}_{best}$ and $\varepsilon$.
Given that the shift of the best fitting model, for CIFAR-10, is between $\varepsilon=10\%$ and $\varepsilon=20\%$, we propose down here a study including also the new noise levels $\varepsilon={12.5\%, 15\%, 17.5\%}$, following the same experimental setup we use and detailed in Appendix~\ref{Appendix: details on the learning strategies} (ResNet-18).
\begin{table}[h]
    \centering
    \begin{tabular}{c c c c}
    \toprule
    \bf Noise rate $\varepsilon$ [\%] & \bf Best Accuracy [\%] & \bf Sparsity [\%] & \bf Best Accuracy Phase \\
    \midrule
    10 & 89.66 & 86.58 & Light Phase (I) \\
    12.5 & 87.57 & 86.58 & Light Phase (I) \\
    15 & 86.98 & 99.81 & Sweet Phase (III) \\
    17.5 & 86.29 & 99.84 & Sweet Phase (III) \\
    20 & 85.72 & 99.81 & Sweet Phase (III) \\
    50 & 77.12 & 99.88 & Sweet Phase (III) \\
    \bottomrule
    \end{tabular}
    \caption{Correlation between noise and $\boldsymbol{w}_{best}$}
    \label{tab:Correlation noise and wbest}
\end{table}

Empirically, we observe a correlation between the best model and the noise rate in the dataset: on CIFAR-10, $\boldsymbol{w}_{best}$  is located in the Light Phase (I) for small noise rate (i.e. $< 15\%$). Once, the noise rate exceeds $15\%$, $\boldsymbol{w}_{best}$ is consistently found in the Sweet Phase (III).
\section{Study employing structured $\ell_1$-pruning}
\label{Appendix: structured sparsity}

We mainly consider unstructured pruning as it is also the one chosen pruning strategy to evidence SDD in~\cite{SparseDoubleDescent}. However, we propose here below a study employing structured $\ell_1$-pruning on CIFAR-10 with 
$\varepsilon=50\%$, in the same experimental setup detailed in Appendix B (but with ($\delta=1$).

\begin{table}[H]
    \centering
    \begin{tabular}{c c c}
        \toprule
        \bf \# Filters & \bf Vanilla test accuracy [\%] & \bf Distillation from pruned teacher test accuracy [\%]\\
        \midrule
        512 & 66.80 & 73.08\\
        256 & 67.10 & 71.54\\
        128 & 68.43 & 70.86\\
        64 & 66.88 & 67.96 \\
        32 & 65.20 & 65.70 \\
        \bottomrule
    \end{tabular}
    \caption{Study employing structured $\ell_1$-pruning.}
    \label{tab:structured pruning}
\end{table}

Empirically, we still observe similar behavior as with unstructured pruning: for the vanilla model there is an evident non-monotonic behavior as the number of filters is reduced in the convolutional block, while with our distillation approach the trend is monotonic, and the performance is consistently higher than with the vanilla approach.
\section{Re-enabling early-stop criteria}
\label{Appendix:early-stop}
We present here an overall approach enabling-back the use of early-stop criteria jointly with the entropy for a given model $\mathcal{M}$ in Alg.~\ref{Algo_early_stop}. Indeed, as highlighted in Fig~\ref{fig:phases}, the entropy stays stationary and then decreases when the model enters the classical regime. Using traditional early criteria starting from this regime can save training computation as the pruning/training process is stopped when the performance decreases.

\begin{algorithm}
\caption{Re-enabling early-stop.}
\label{Algo_early_stop}
\begin{algorithmic}[1]
    \FUNCTION{Early-stopping($\boldsymbol{w}^\mathcal{M}$, $\Xi$, $\zeta$, $\mathcal{T}_E$, $\mathcal{T}_A$)}{}
    \STATE $\boldsymbol{w}^\mathcal{M}\gets $Train($\boldsymbol{w}^\mathcal{M}$, $\Xi_{train}$) \alglinelabel{2_line:train0}
    \STATE best\_acc $\gets$ Evaluate($\zeta_{\text{current}}$, $\Xi_{val}$)
    \STATE $\eta_0 \gets$ Entropy($\boldsymbol{w}^\mathcal{M}$, $\zeta_{\text{current}}$, $\Xi_{train}$)
    \STATE $\eta_{\text{current}}\gets \eta_0$
    \STATE $\zeta_{\text{current}}\gets \zeta$
    \WHILE{$\eta_{\text{current}}$ > $\eta_0 \times \mathcal{T}_E$} \alglinelabel{2_line:while1}
        \STATE $\boldsymbol{w}^\mathcal{M}$, this\_acc $\gets$ PPTE($\boldsymbol{w}^\mathcal{M}$, $\zeta_{\text{current}}$, $\Xi_{train}$, $\Xi_{val}$) \alglinelabel{2_line:function1}
        \STATE best\_acc $\gets$ $\max$(this\_acc, best\_acc)
         \STATE $\eta_{\text{current}} \gets$ Entropy($\boldsymbol{w}^\mathcal{M}$, $\zeta_{\text{current}}$, $\Xi_{train}$)
        \STATE $\zeta_{\text{current}} \gets 1 - (1-\zeta_{\text{current}}) (1-\zeta)$
    \ENDWHILE
    \WHILE{this\_acc$>$best\_acc $\times \mathcal{T}_A$} \alglinelabel{2_line:while2}
        \STATE $\boldsymbol{w}^\mathcal{M}$, this\_acc $\gets$ PPTE($\boldsymbol{w}^\mathcal{M}$, $\zeta_{\text{current}}$, $\Xi_{train}$, $\Xi_{val}$) \alglinelabel{2_line:function2}
        \STATE $\zeta_{\text{current}} \gets 1 - (1-\zeta_{\text{current}}) (1-\zeta)$
    \ENDWHILE
    \STATE \textbf{return} $\boldsymbol{w}^\mathcal{M}$ \alglinelabel{2_line:return}
    \ENDFUNCTION
\end{algorithmic}
\end{algorithm}

The first step is to train the dense model (line~\ref{2_line:train0}).
While the entropy $\eta$ calculated on the training set $\Xi_{train}$ remains stationary, i.e. when $\eta_{\text{current}} > \eta_0 \times \mathcal{T}_E$ (line~\ref{2_line:while1}), where $\mathcal{T}_E$ represents a threshold (e.g. $80\%$) and $\eta_0$ the entropy of the model after the first training, the model is iteratively pruned, perturbed and re-trained on $\Xi_{train}$ (line~\ref{2_line:function1}) using the function PPTE defined in Appendix~\ref{Appendix: details on the learning strategies}.\\
Once $\eta_{\text{current}} < \eta_0 \times \mathcal{T}_E$, i.e. when the entropy is decreasing, we re-enable an early-stop criterion. 
Until the current performance this\_acc on the validation set is lower than best\_acc $\times \mathcal{T}_A$,  where $\mathcal{T}_A$ represents a threshold (e.g. $80\%$), we continue to prune, perturb and re-train on $\Xi_{train}$ (line~\ref{2_line:function2}).  
When the performance is below best\_acc $\times \mathcal{T}_A$, the algorithm returns the model parameters $\boldsymbol{w}^\mathcal{M}$ (line~\ref{2_line:return}).

\newpage\section{Experiments on more datasets}
\label{sec:moredatasets}

In this section we will present some results, with training on-the-wild (or in different terms, without injecting noise and using common and standard learning policies), on two main-stream datasets: Flowers-102 and ImageNet-1k. 
Moreover, as synthetic noise has clean structures which greatly enabled statistical analyses but often fails to model the real-world noise patterns, we also conducted experiments on CIFAR-100N, a dataset presented by~\cite{wei2022learning}, which is formed with CIFAR-100 training dataset equipped with human-annotated real-world noisy labels collected from Amazon Mechanical Turk.

\subsection{CIFAR-100N} 

For CIFAR-100N, we used the same learning policy as for CIFAR-10 for the VGG-like architectures as reported in Tab.~\ref{tab: Learning strategies}. We have employed a ResNet-18 as teacher model while a VGG-like model with $\gamma=5$ and $\delta=5$ as student.
\begin{figure}[ht]
    \begin{subfigure}{0.5\textwidth}
        \includegraphics[width=\textwidth]{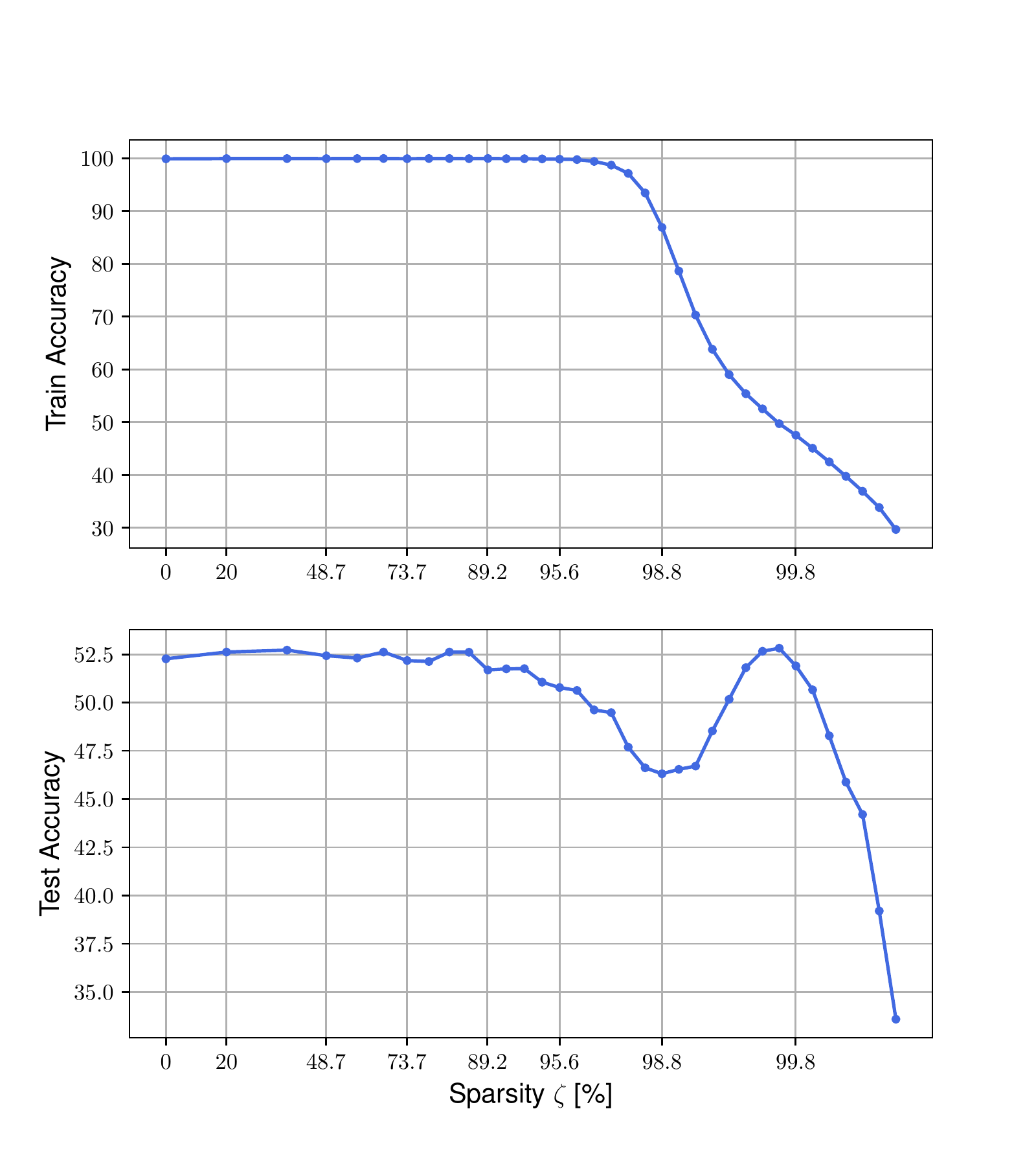}
        \caption{~}
        \label{fig:CIFAR-100N_Teacher}
    \end{subfigure}
    \begin{subfigure}{0.5\textwidth}
        \includegraphics[width=\textwidth]{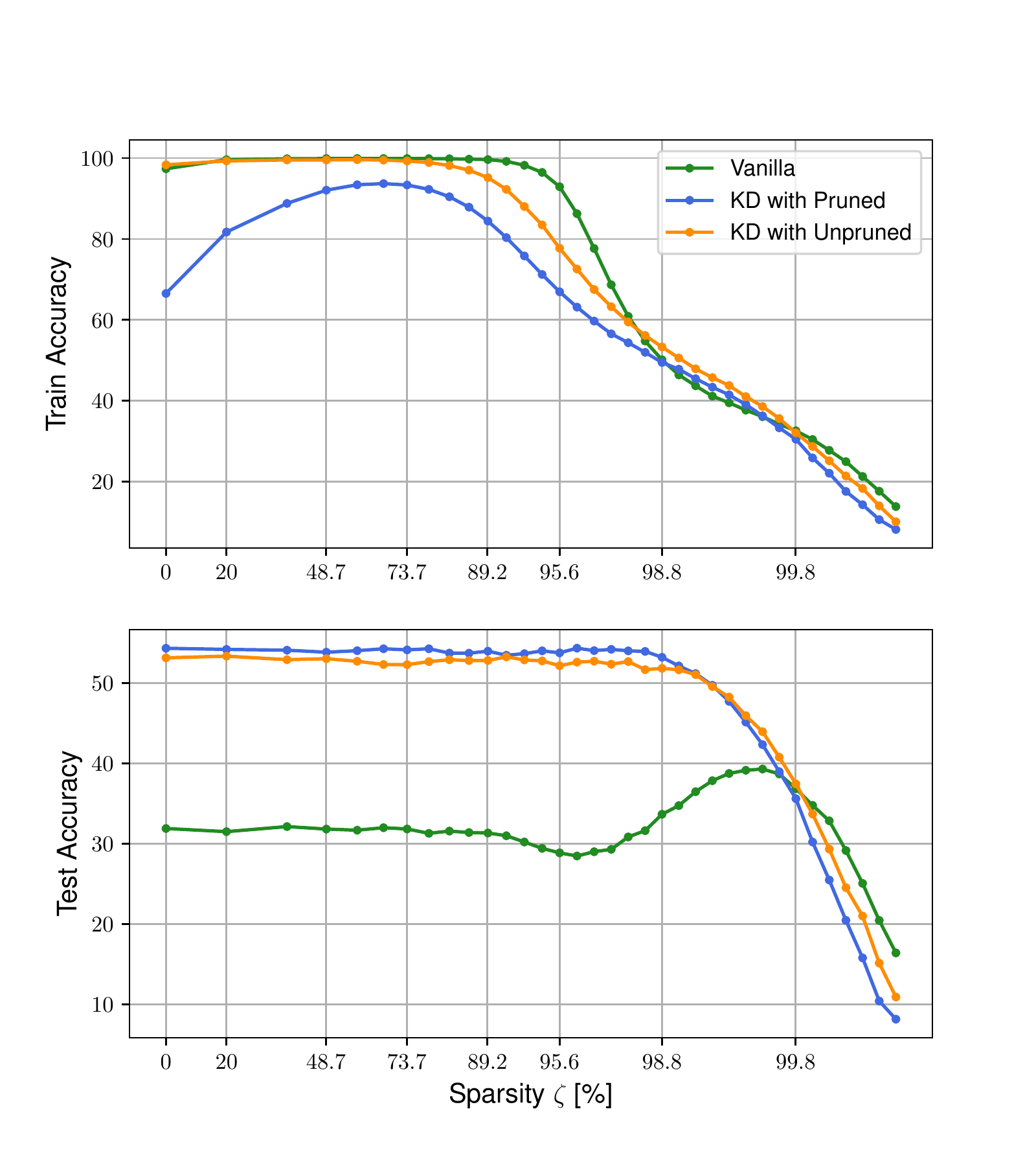}
        \caption{~}
        \label{fig:CIFAR-100N_simple}
    \end{subfigure}
    \caption{Performance on CIFAR-100N. \textbf{Left.} ResNet-18 \textbf{Right.} VGG-like model}
    \label{fig:CIFAR-100N}
\end{figure}

In Fig.~\ref{fig:CIFAR-100N} we report the distillation results on CIFAR-100N. Like for CIFAR-10 and CIFAR-100, as already reported in Sec.~\ref{subsec : Relationships between DD and sparse DD}, we consistently observe that, when sparsity increases, the student model, trained in a vanilla setup, exhibits the sparse double descent phenomenon. However, for the same architecture, we persistently notice that employing KD within the framework proposed in Sec.~\ref{sec: KD avoid DD}, whether the teacher is pruned or not (i.e. dense),  the performance is enhanced and exhibits a monotonic behavior: the sparse double descent is dodged.\\
 We also note that, in high sparsity regimes, the performance of the student model, trained within the KD framework becomes marginally below the vanilla setup. This behaviour can be explained by the fact that $\alpha$ and $\tau$ were not tuned for this dataset. 
 
\subsection{Flowers-102}

\begin{figure}[ht]
    \begin{subfigure}{0.5\textwidth}
        \includegraphics[width=\textwidth]{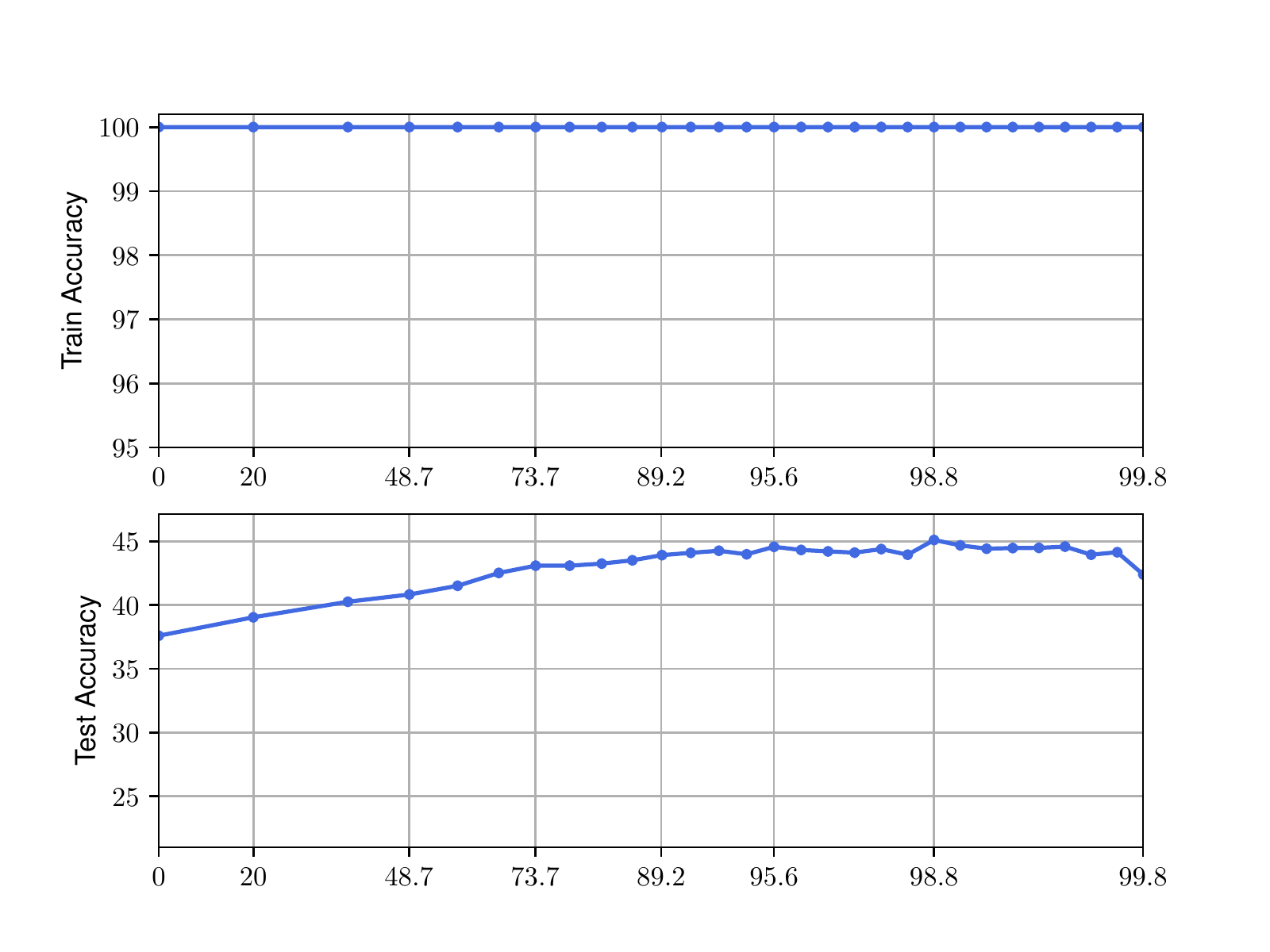}
        \caption{~}
        \label{fig:Flowers_Teacher}
    \end{subfigure}
    \begin{subfigure}{0.5\textwidth}
        \includegraphics[width=\textwidth]{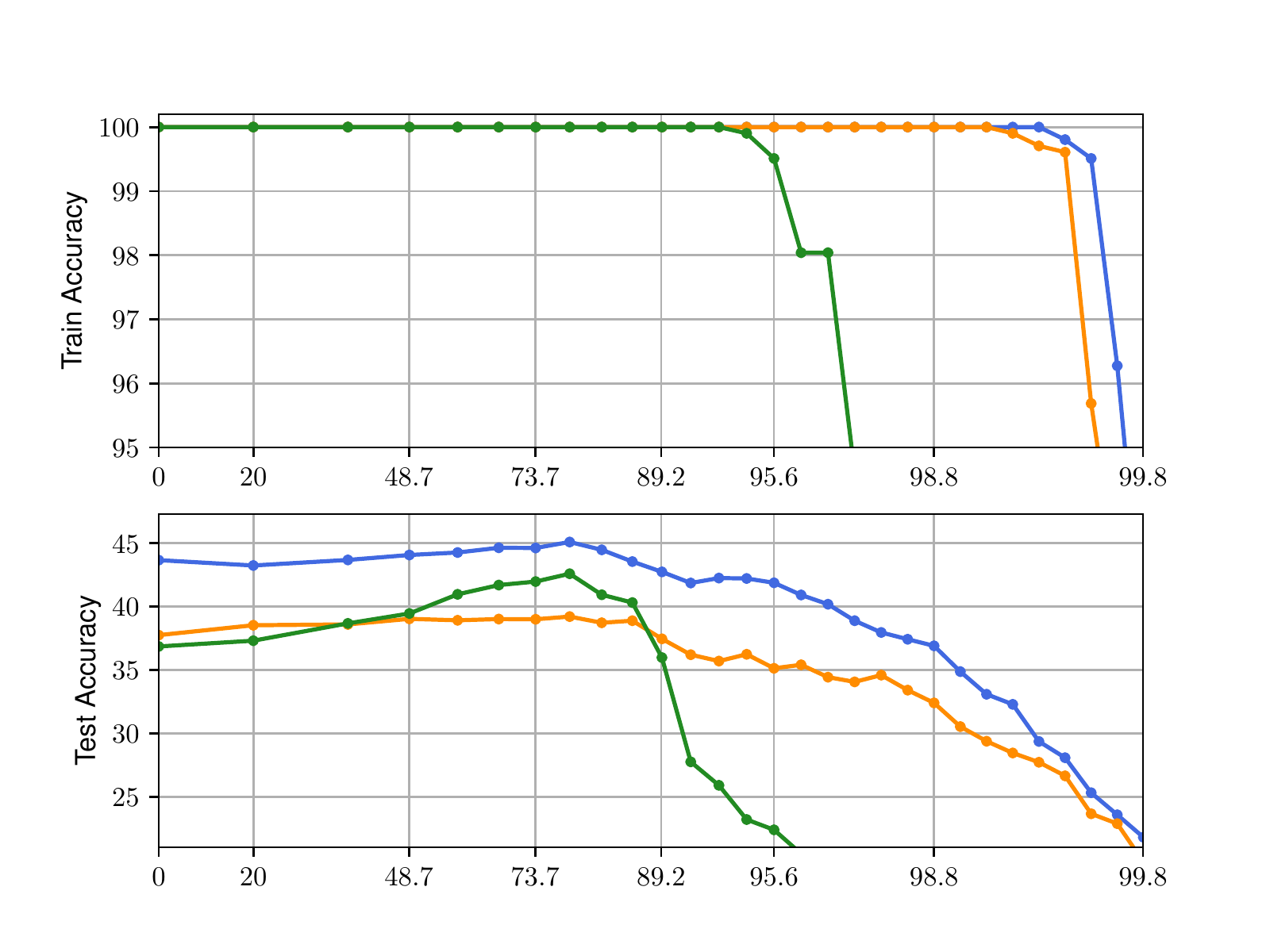}
        \caption{~}
        \label{fig:Flowers_simple}
    \end{subfigure}
    \caption{Performance on Flowers-102. \textbf{Left.} ResNet-18 \textbf{Right.} VGG-like model}
    \label{fig:Flowers}
\end{figure}

For Flowers-102, we have employed exactly the same strategy as for CIFAR-10 for the VGG-like architectures, as reported in Tab.~\ref{tab: Learning strategies}. We have employed for our experiment a ResNet-18 as teacher model while a VGG-like model with $\gamma=5$ and $\delta=4$ as student. Given the reduced size of the train set (consisting in 1020 samples, 10 per class), evidently, over-fitting is in general very easy for a sufficiently-large model. In order to further enhance this effect, we have decided not to use any dataset augmentation strategy. According to the results reported in Fig.~\ref{fig:Flowers}, we clearly observe that, despite an evident overfit from both teacher and student, no double descent is visible, although the generalization gap is huge.\footnote{Employing transfer learning strategies it is in general possible to achieve extremely high performance on this performance, above 95\%, on the test set. However, this is not the scope of the paper.} In this case, we hypothesize that the noise is irrelevant at the dataset's scale, and the model simply lacks the proper priors to learn the right set of features.

\subsection{ImageNet}
For ImageNet, we have employed the standard learning policy, consisting in SGD with lr decayed at epoch milestones 30 and 60, trained for 90 epochs with initial learning rate 0.1 and momentum 0.9, using batchsize 128. We have employed for our experiment a ResNet-50 as teacher model while a ResNet-18 as student. The result is reported in Tab.~\ref{tab:ImageNet}. Unsurprisingly, we do not observe a sparse double descent for any of the considered configurations: it is known that even ResNet-50 is an under-parametrized model with respect to ImageNet, and evidently we are already in the collapsed phase.
\begin{table}[H]
    \resizebox{\textwidth}{!}{
    \centering
    \begin{tabular}{c c c c }
        \toprule
        $\zeta$ & ResNet-18 (Vanilla) & ResNet-18 (KD with a dense ResNet-50) & ResNet-18 (KD with a 50\% pruned ResNet-50)\\
        \midrule
        0 &  68.89 & 69.57  & 69.75 \\
        0.5 & 69.28  & 69.82  & 69.63\\
        0.75 & 68.98  & 69.03  & 68.93\\
        0.875 & 67.50  & 66.04  & 66.08\\
        \bottomrule
    \end{tabular}
    }
    \caption{Test accuracy of ResNet-18 on ImageNet.}
    \label{tab:ImageNet}
\end{table}

\newpage\section{Various visualizations for varying width \& depth experiments}
\label{sec:appWH}


\subsection{Study on the width for CIFAR-10 with $\varepsilon=50\%$}

\begin{figure}[H]
    \centering
    \begin{subfigure}[H]{0.45\textwidth}
        \includegraphics[width=1.0\linewidth]{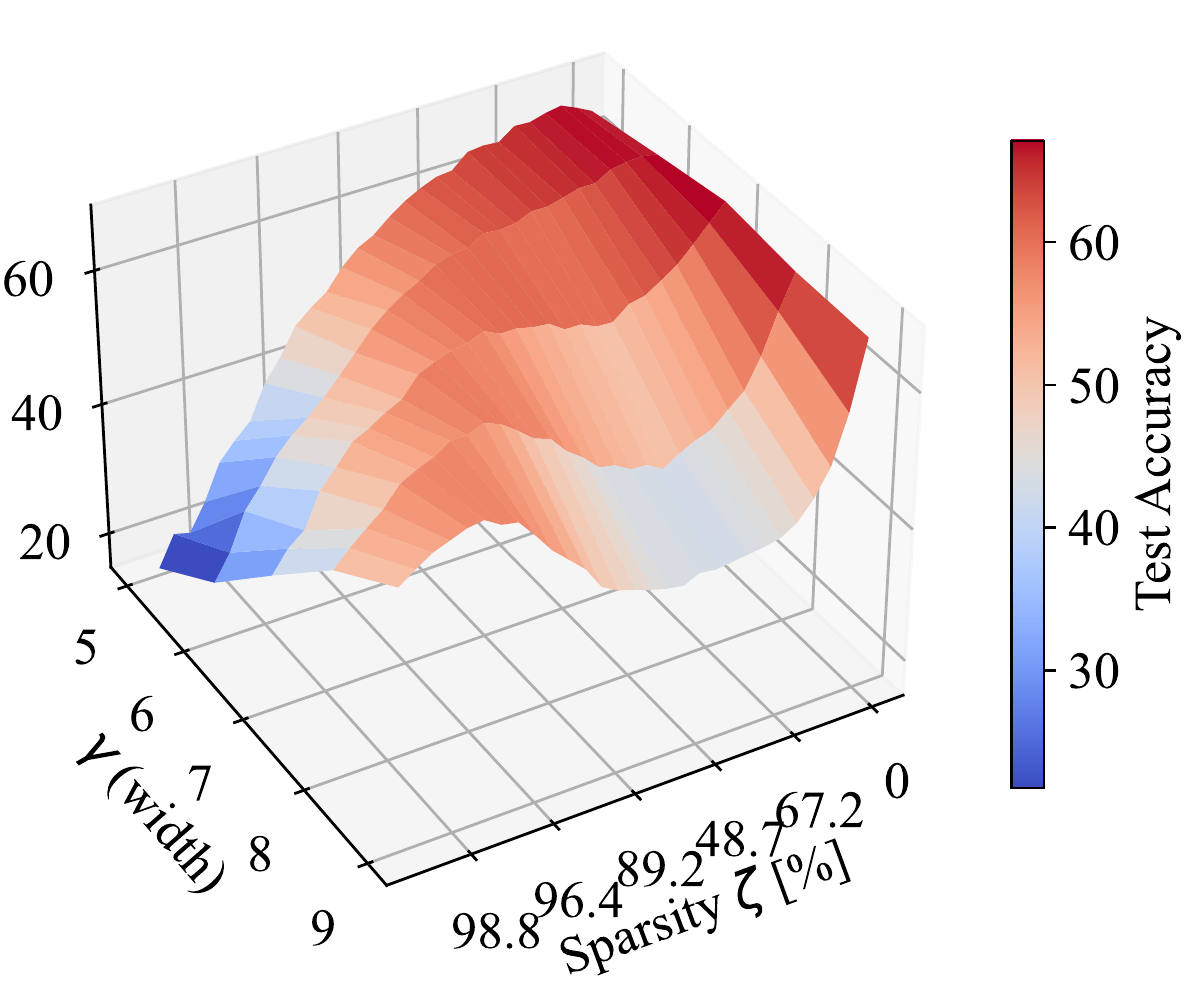}
    \end{subfigure}
    \begin{subfigure}[H]{0.45\textwidth}
        \includegraphics[width=1.0\linewidth]{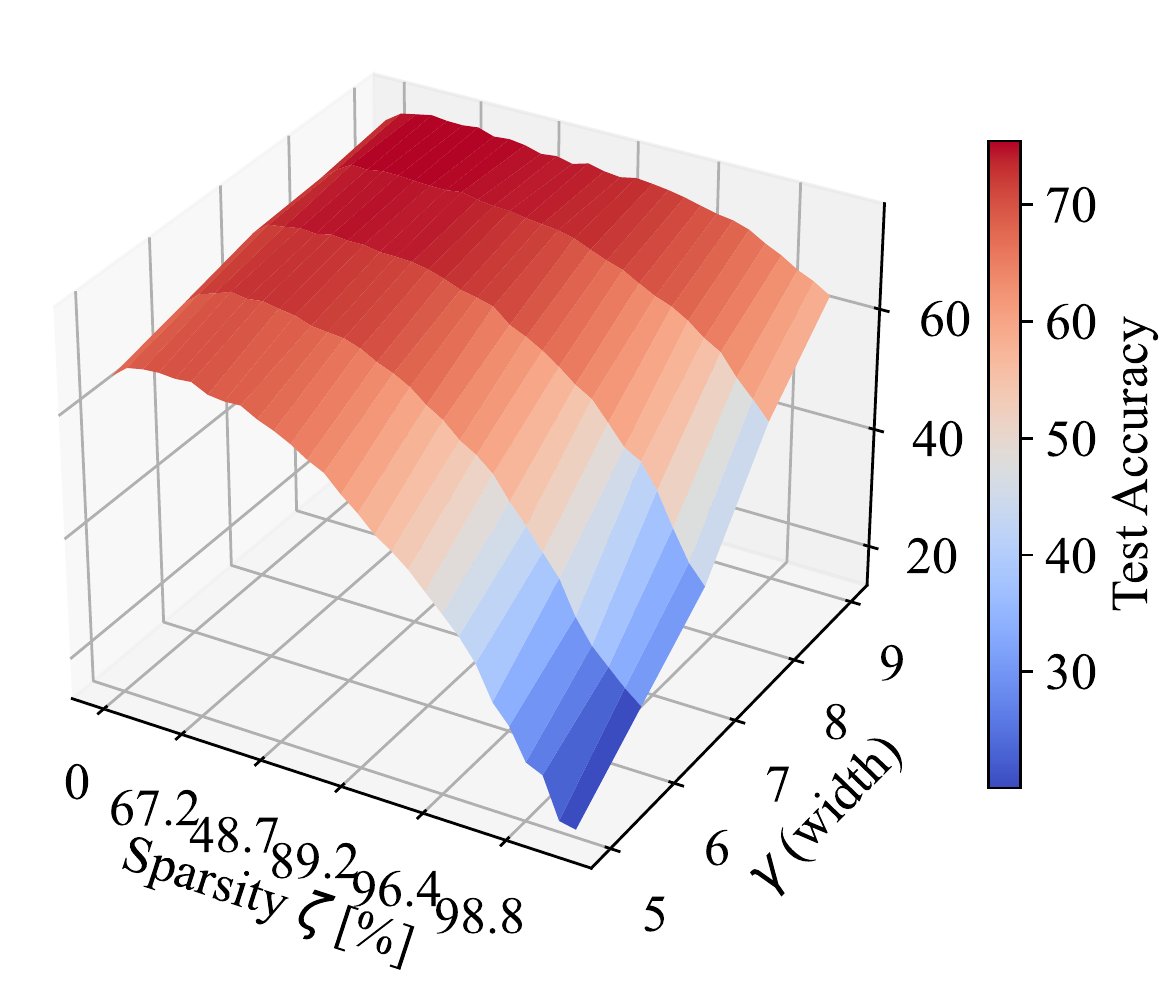}
    \end{subfigure}
    \caption{Test accuracy of the VGG-like varying $\gamma$ on CIFAR-10 with $\varepsilon=50\%$.
    \textbf{Left:} Vanilla Training. \textbf{Right:} Our proposed method.}
    \label{fig:3D_CIFAR-10_50_Width}
\end{figure}

\subsection{Study on the Depth for CIFAR-10 with $\varepsilon=50\%$}

\begin{figure}[H]
    \centering
    \begin{subfigure}[H]{0.45\textwidth}
        \includegraphics[width=1.0\linewidth]{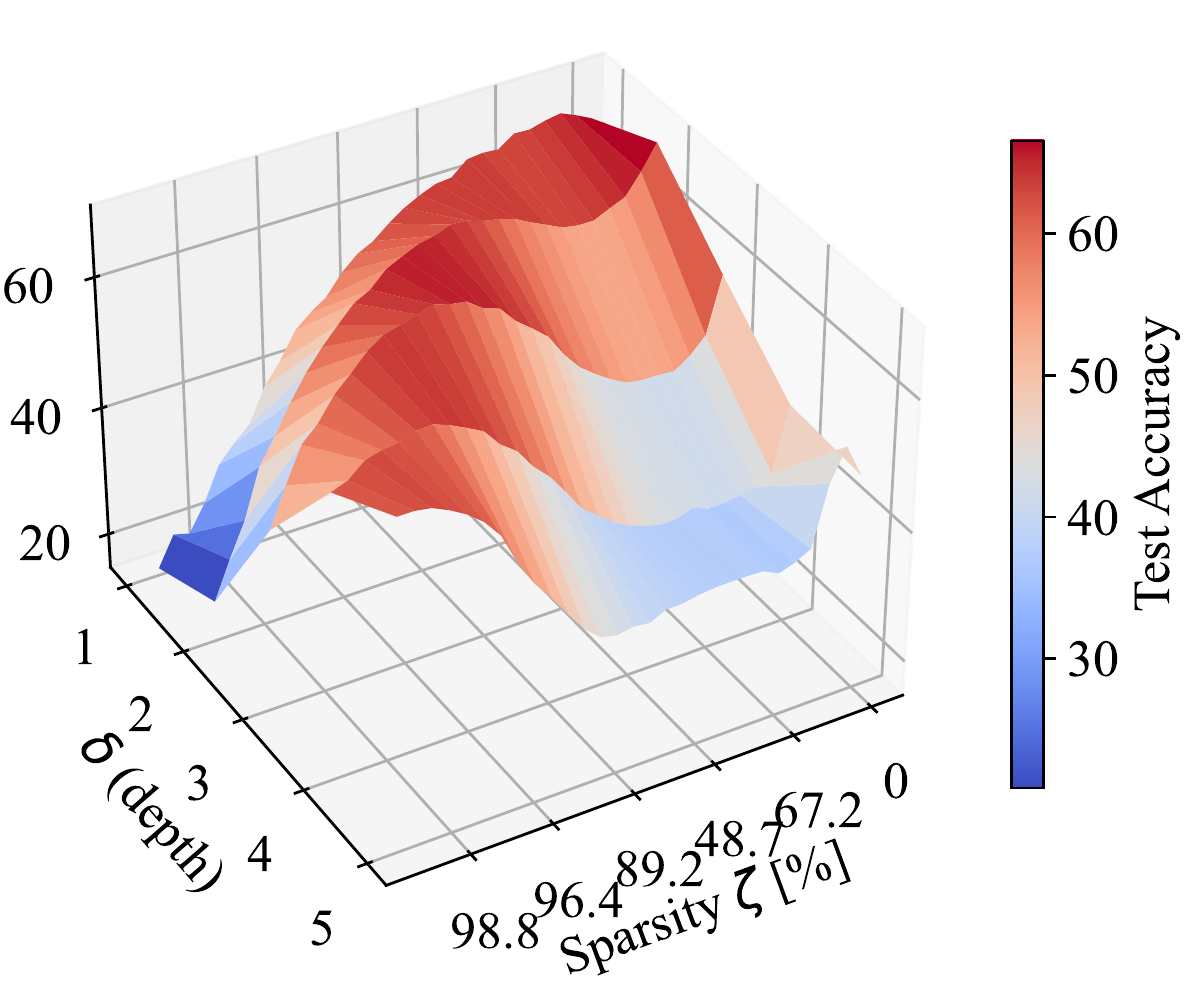}
    \end{subfigure}
    \begin{subfigure}[H]{0.45\textwidth}
        \includegraphics[width=1.0\linewidth]{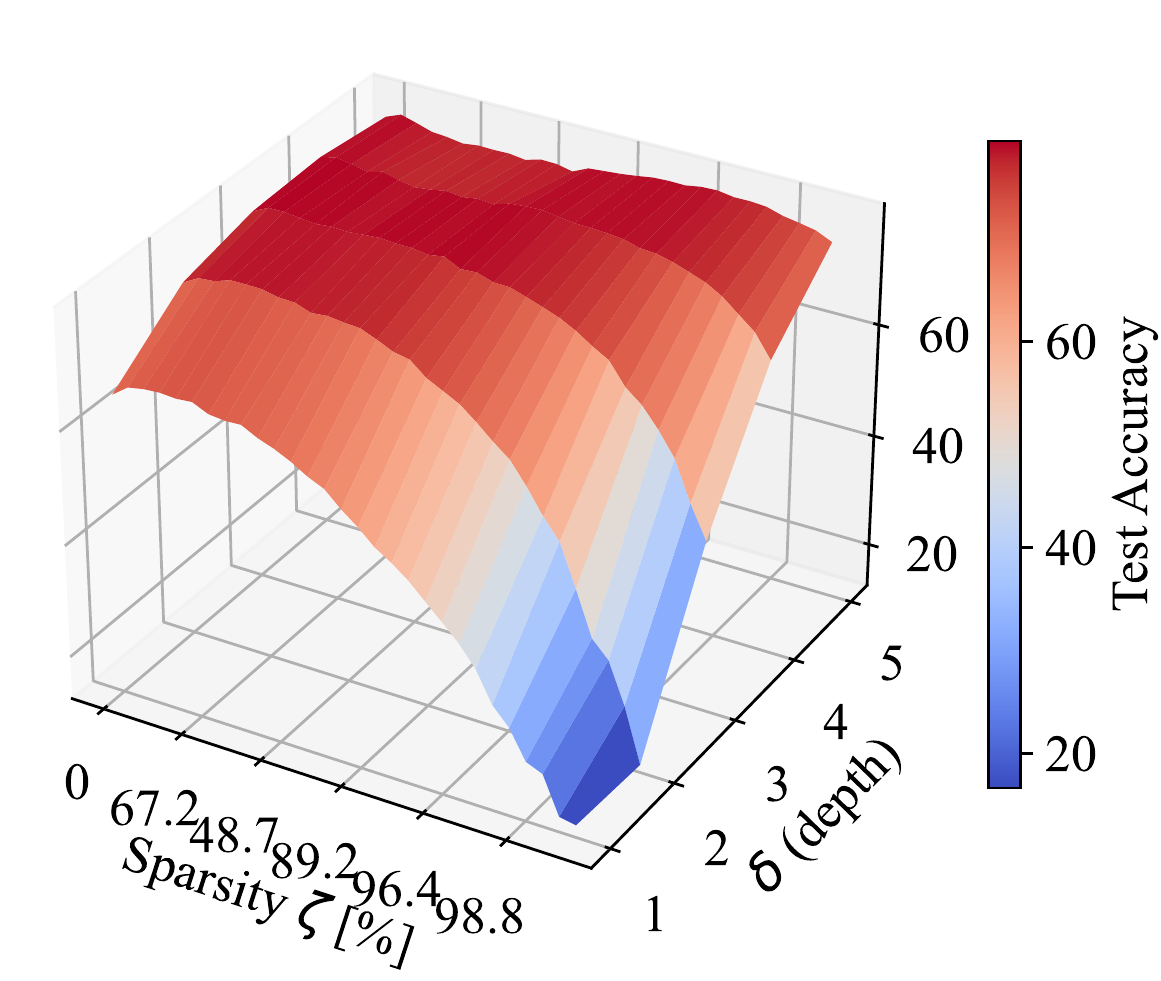}
    \end{subfigure}
    \caption{Test accuracy of the VGG-like varying $\delta$ on CIFAR-10 with $\varepsilon=50\%$.
    \textbf{Left:} Vanilla Training. \textbf{Right:} Our proposed method.}
    \label{fig:3D_CIFAR-10_50_Depth}
\end{figure}


\subsection{Study on the width for CIFAR-10 with $\varepsilon=20\%$}

\begin{figure}[H]
    \centering
    \begin{subfigure}[H]{0.45\textwidth}
        \includegraphics[width=1.0\linewidth]{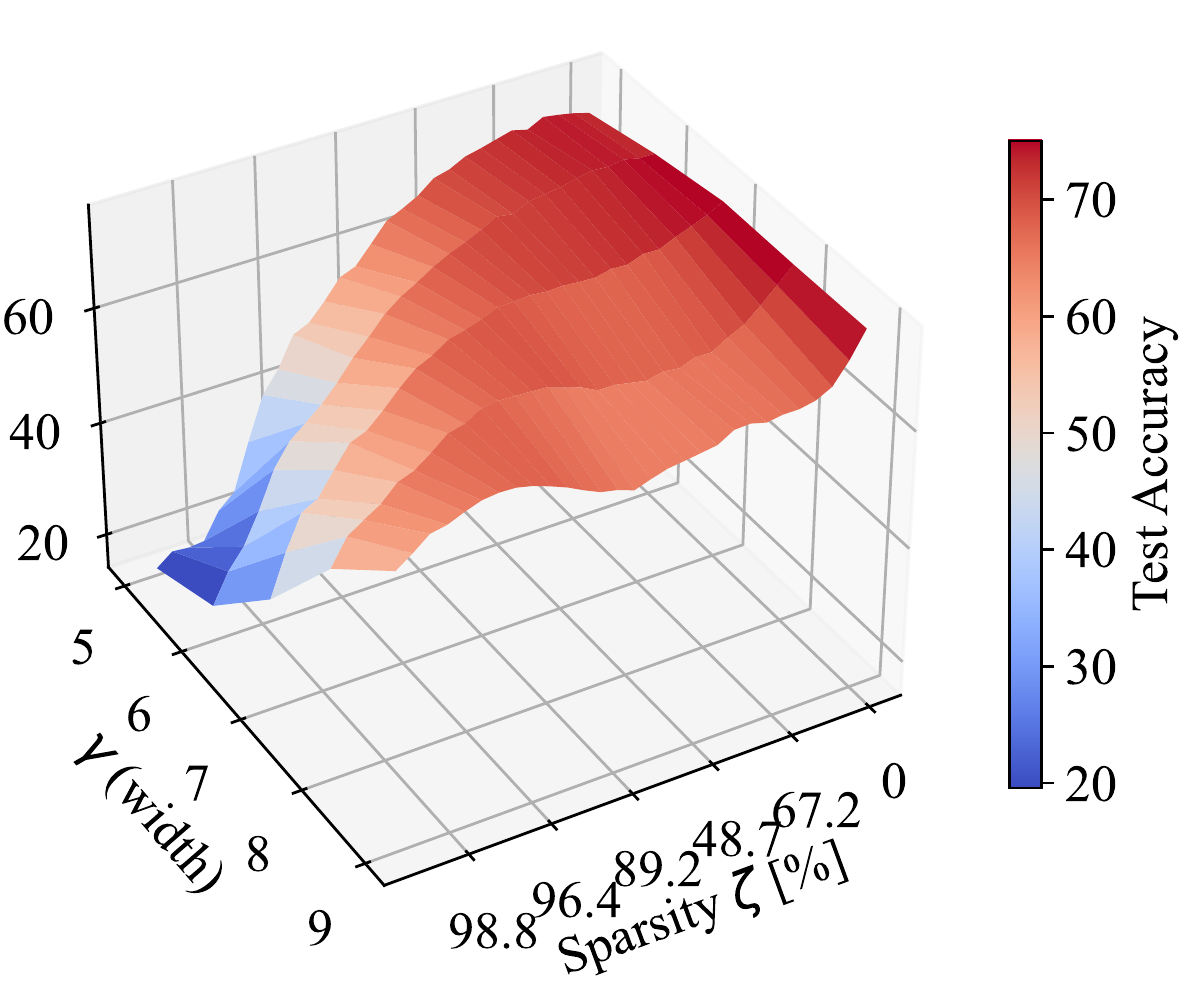}
    \end{subfigure}
    \begin{subfigure}[H]{0.45\textwidth}
        \includegraphics[width=1.0\linewidth]{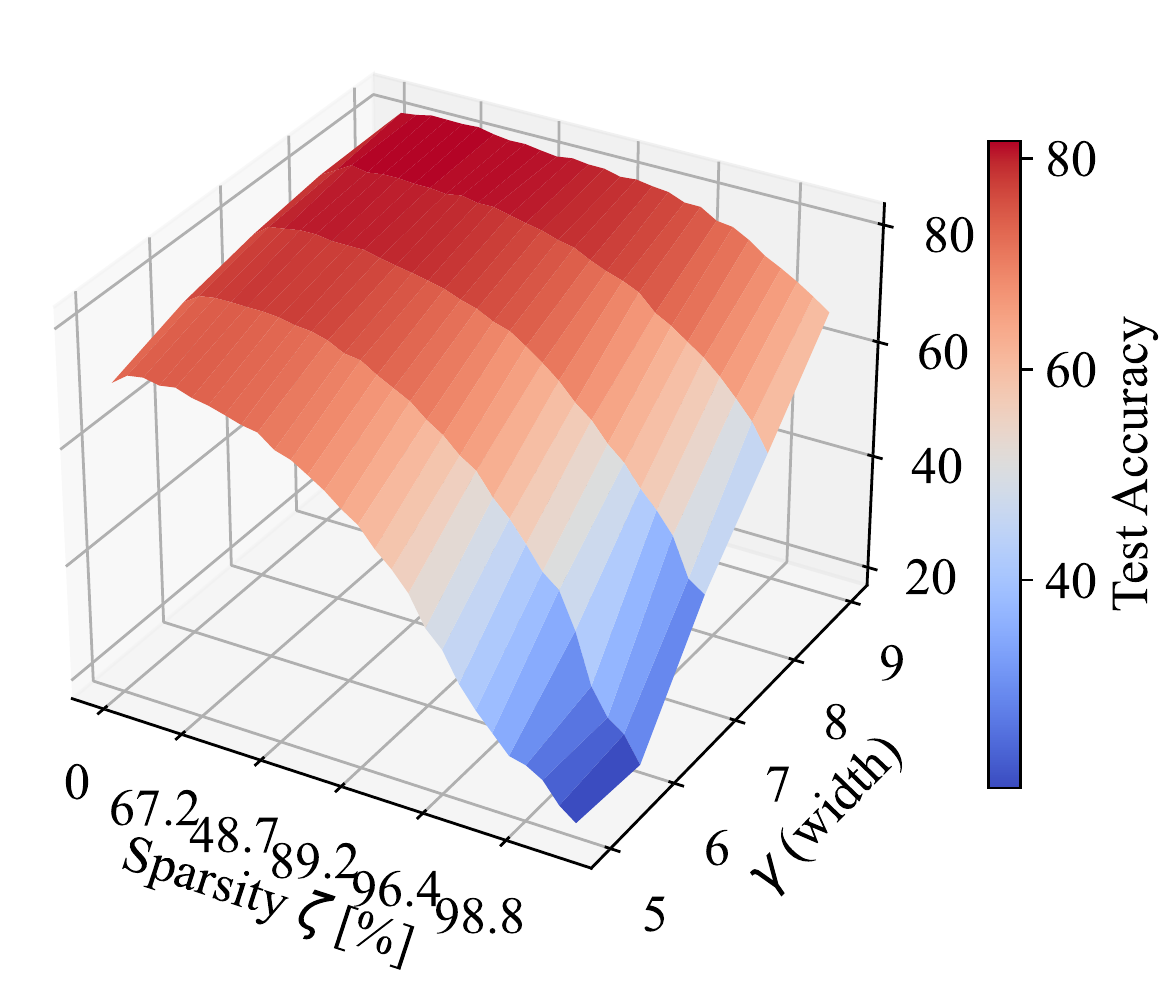}
    \end{subfigure}
    \caption{Test accuracy of the VGG-like varying $\gamma$ on CIFAR-10 with $\varepsilon=20\%$.
    \textbf{Left:} Vanilla Training. \textbf{Right:} Our proposed method.}
    \label{fig:3D_CIFAR-10_20_Width}
\end{figure}

\subsection{Study on the depth for CIFAR-10 with $\varepsilon=20\%$}

\begin{figure}[H]
    \centering
    \begin{subfigure}[H]{0.45\textwidth}
        \includegraphics[width=1.0\linewidth]{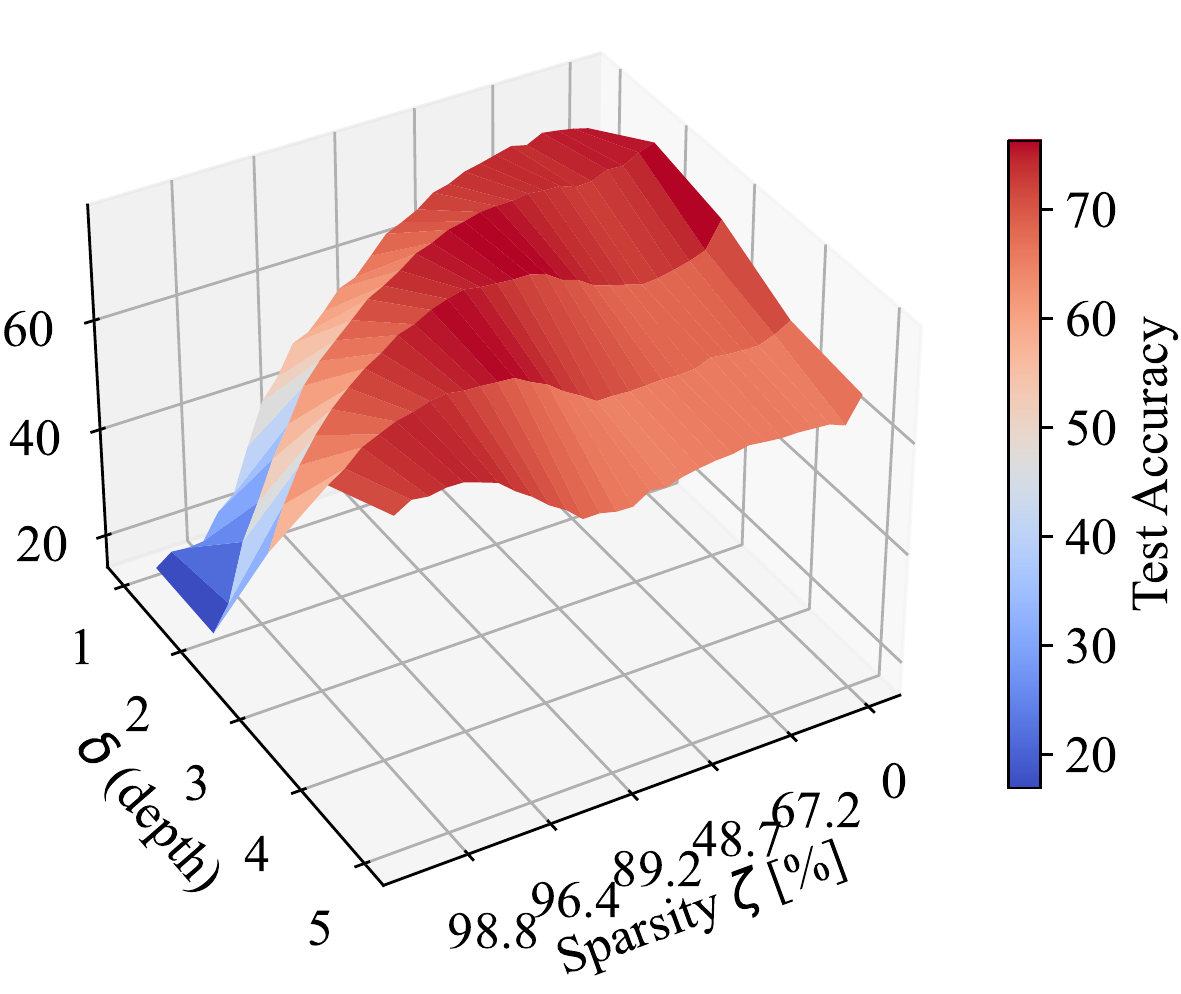}
    \end{subfigure}
    \begin{subfigure}[H]{0.45\textwidth}
        \includegraphics[width=1.0\linewidth]{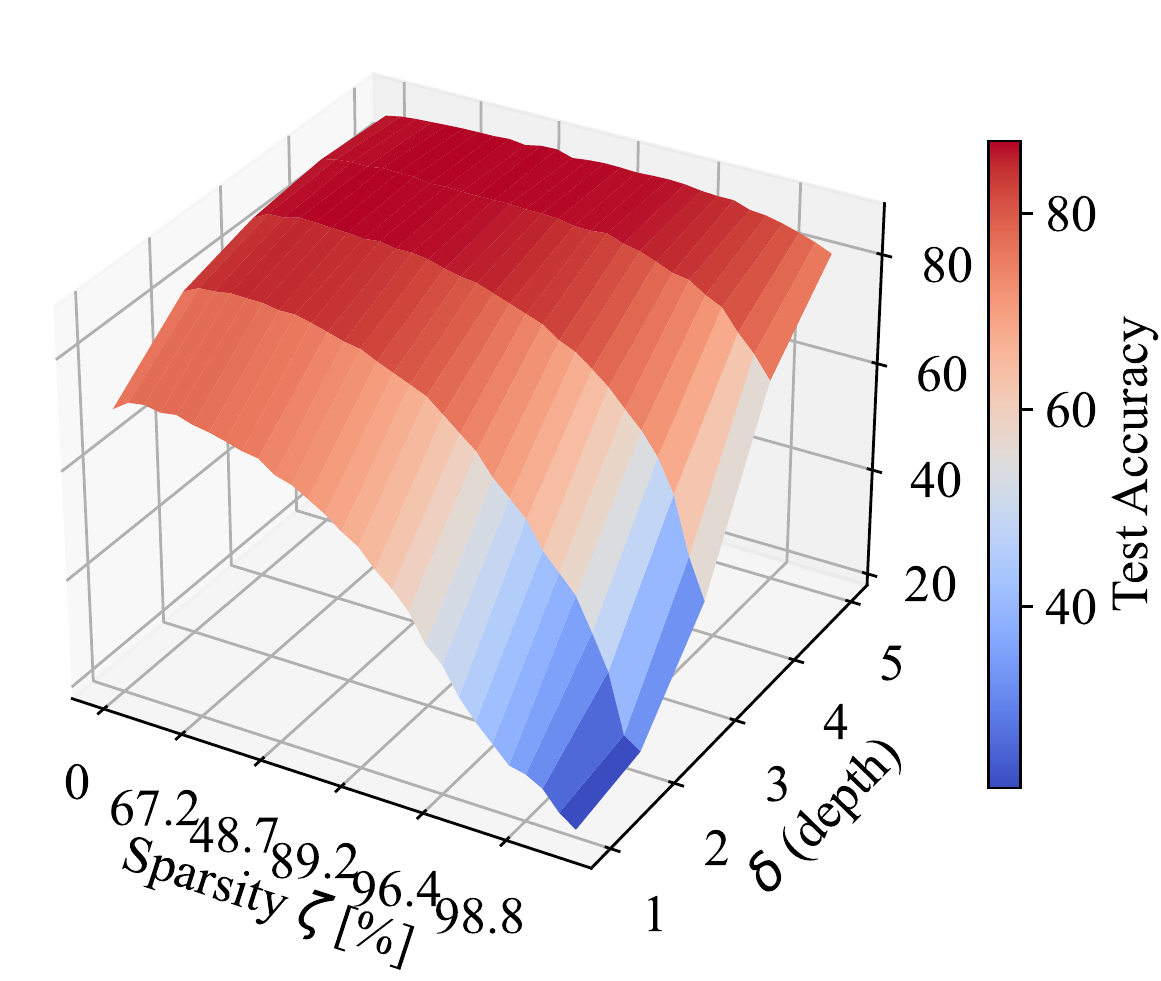}
    \end{subfigure}
    \caption{Test accuracy of the VGG-like varying $\delta$ on CIFAR-10 with $\varepsilon=20\%$.
    \textbf{Left:} Vanilla Training. \textbf{Right:} Our proposed method.}
    \label{fig:3D_CIFAR-10_20_Depth}
\end{figure}

\subsection{Study on the width for CIFAR-10 with $\varepsilon=10\%$}

\begin{figure}[H]
    \centering
    \begin{subfigure}[H]{0.45\textwidth}
        \includegraphics[width=1.0\linewidth]{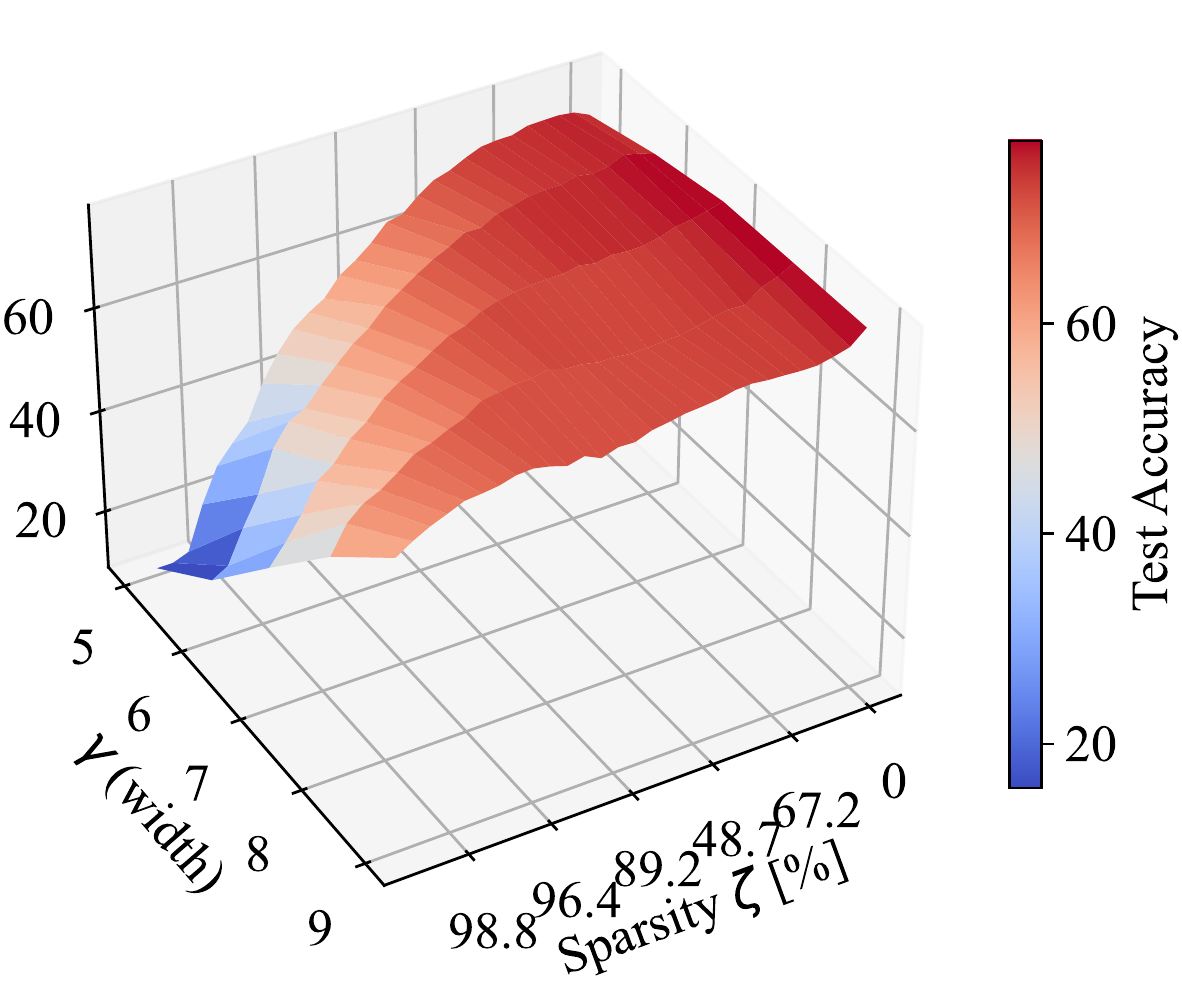}
    \end{subfigure}
    \begin{subfigure}[H]{0.45\textwidth}
        \includegraphics[width=1.0\linewidth]{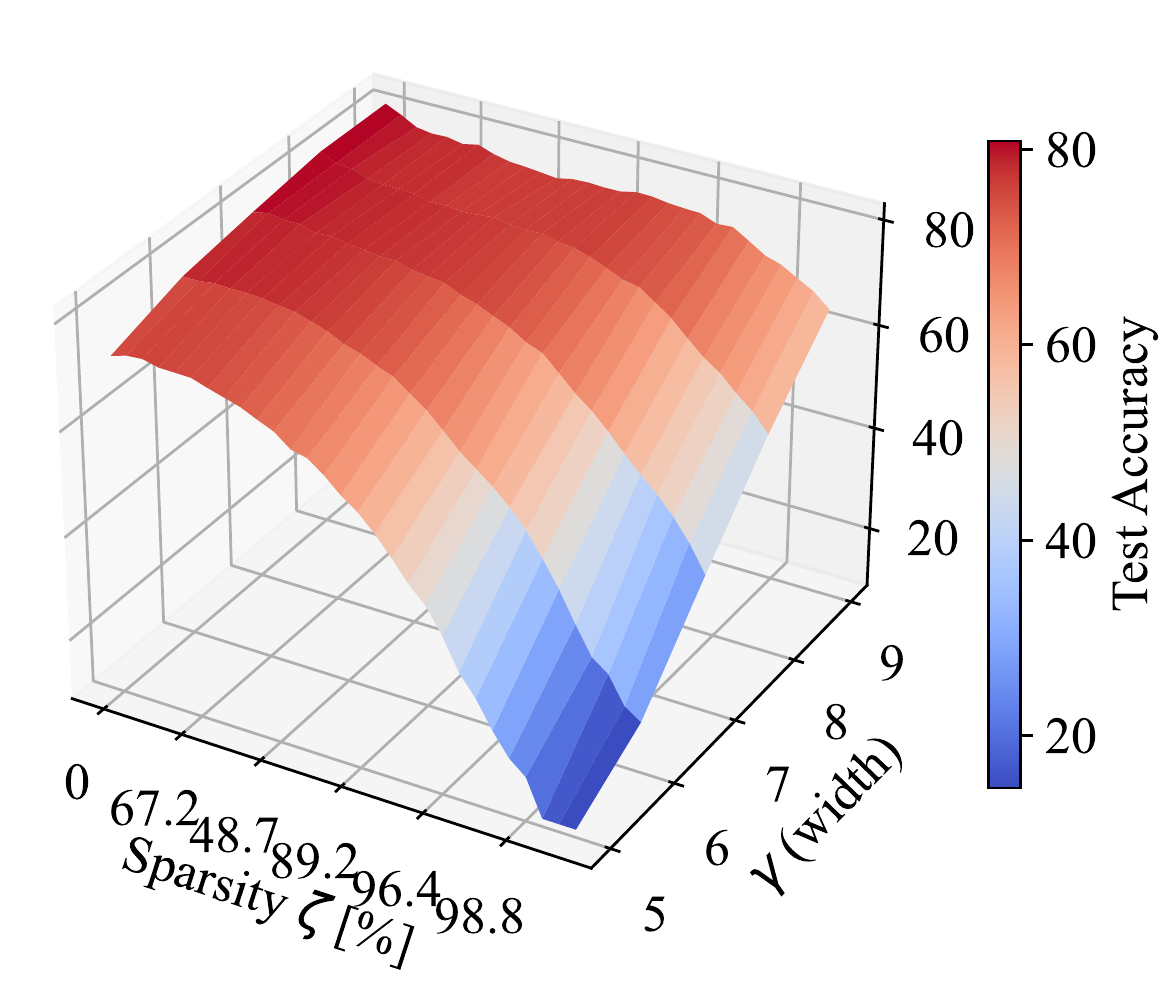}
    \end{subfigure}
    \caption{Test accuracy of the VGG-like varying $\gamma$ on CIFAR-10 with $\varepsilon=10\%$.
    \textbf{Left:} Vanilla Training. \textbf{Right:} Our proposed method.}
    \label{fig:3D_CIFAR-10_10_Width}
\end{figure}

\subsection{Study on the depth for CIFAR-10 with $\varepsilon=10\%$}
\begin{figure}[H]
    \centering
    \begin{subfigure}[H]{0.45\textwidth}
        \includegraphics[width=1.0\linewidth]{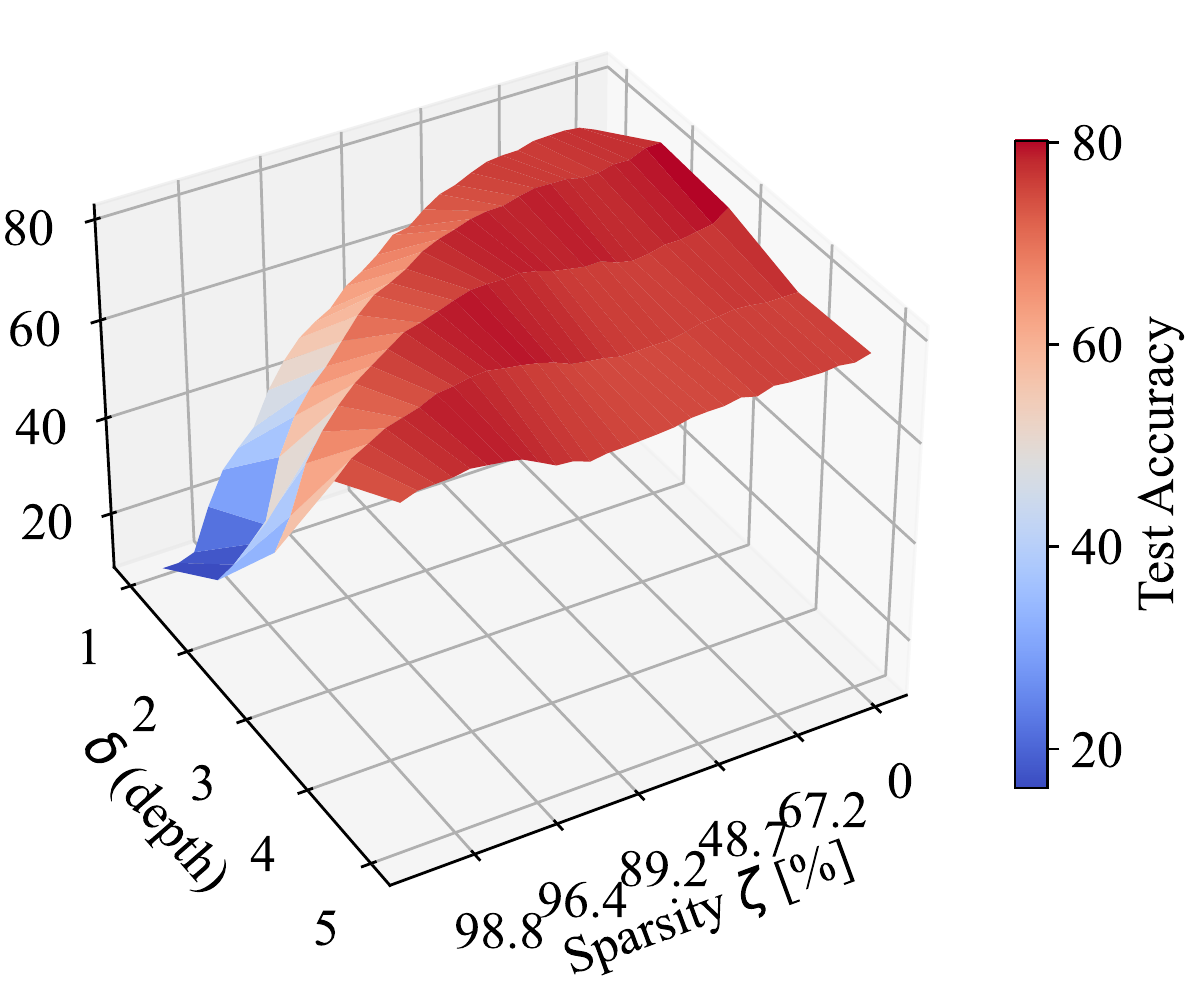}
    \end{subfigure}
    \begin{subfigure}[H]{0.45\textwidth}
        \includegraphics[width=1.0\linewidth]{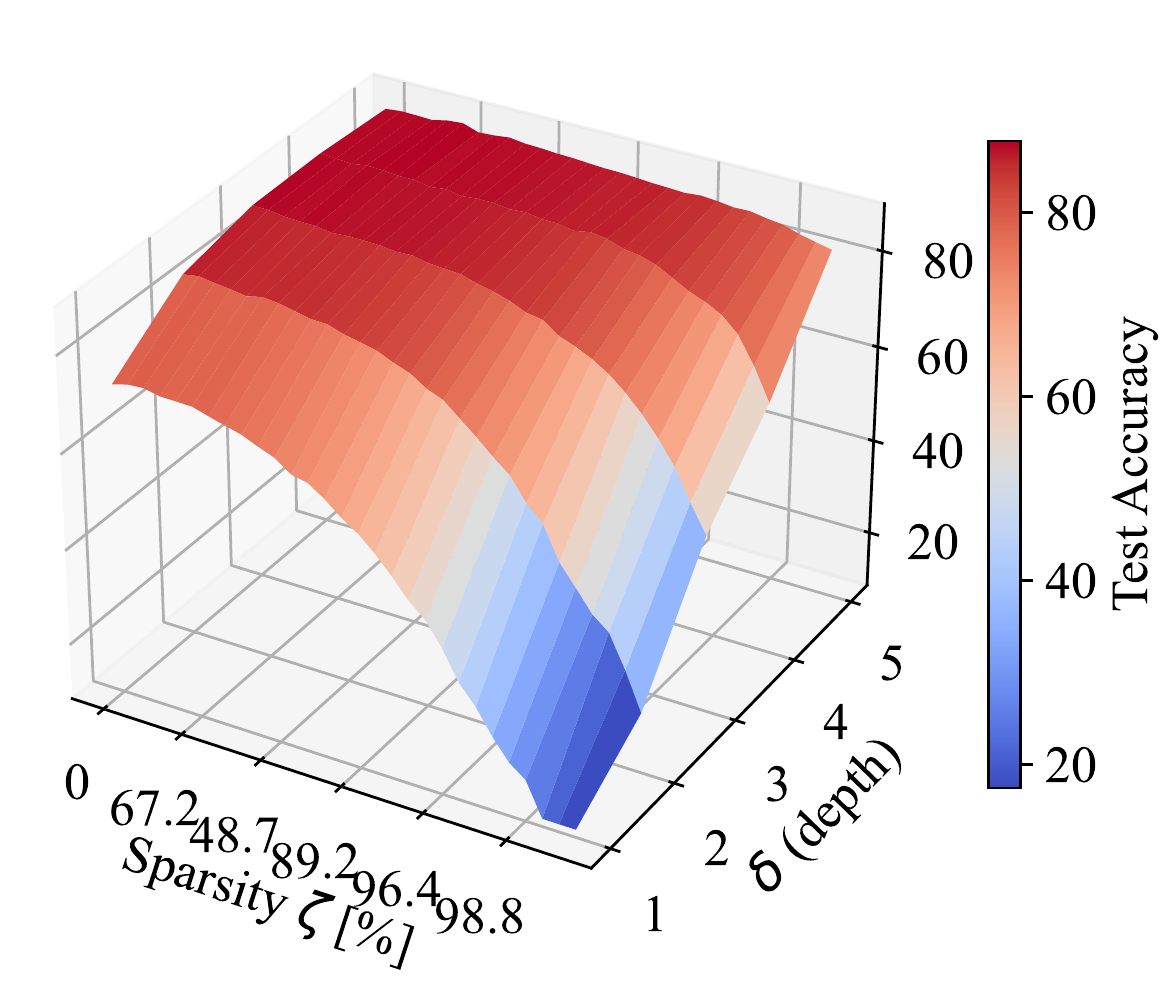}
    \end{subfigure}
    \caption{Test accuracy of the VGG-like varying $\delta$ on CIFAR-10 with $\varepsilon=10\%$.
    \textbf{Left:} Vanilla Training. \textbf{Right:} Our proposed method.}
    \label{fig:3D_CIFAR-10_10_Depth}
\end{figure}

\end{document}